\pdfoutput=1
\documentclass[twocolumn]{svjour3}       
\smartqed  
\usepackage{graphicx}
\usepackage{epsfig}
\usepackage{latexsym}
\usepackage{float}
\usepackage{times}
\usepackage{multirow}
\usepackage{url}
\usepackage{subfig}
\usepackage{threeparttable}
\usepackage{color}
\usepackage[misc]{ifsym}
\usepackage{natbib}
\usepackage{booktabs}

\usepackage{epsfig,latexsym}
\usepackage{float}
\usepackage{indentfirst}
\usepackage{amsmath}
\usepackage{amssymb}
\usepackage{times}
\usepackage{psfrag}
\usepackage{subfig}
\usepackage{textcomp}
\usepackage{multirow}
\usepackage{multicol}
\usepackage{stfloats}
\usepackage{url}
\usepackage{multirow}
\usepackage{booktabs}
\usepackage{comment}
\usepackage{threeparttable}

\usepackage{bbding}

\usepackage{makecell}
\usepackage{colortbl}
\usepackage{array}
\definecolor{mygray}{gray}{.9}

\usepackage[pagebackref=true,breaklinks=true,letterpaper=true,colorlinks=true,bookmarks=false,linkcolor=black,citecolor=blue,urlcolor=blue,]{hyperref}

\usepackage[linesnumbered,ruled]{algorithm2e}
\usepackage{amsfonts}
\usepackage{bm}
\usepackage{booktabs}

\usepackage[table]{xcolor}

\usepackage{float}
\usepackage{svg}

\usepackage{color}
\usepackage[normalem]{ulem}
\def\etal{{\em et al.~}}
\def \ie{{\em i.e.}}
\def \eg{{\em e.g.}}
\def \etc{{\em etc.~}}

\newcommand{\tabincell}[2]{\begin{tabular}{@{}#1@{}}#2\end{tabular}}
\newcommand{\xma}{\textcolor{black}} 

\newcommand{\mla}{\textcolor{black}}
\newcommand{\ml}{\textcolor{blue}}
\newcommand{\xm}{\textcolor{red}}

\newcommand{\yfa}{\textcolor{black}}

\journalname{}
\UseRawInputEncoding
\begin{document}

\title{Joint Learning of Visual-Audio Saliency Prediction and Sound Source Localization on Multi-face Videos
}

\titlerunning{Minglang Qiao et al.}


\author{Minglang Qiao\textsuperscript{1} \and Yufan Liu\textsuperscript{3} \and Mai Xu\textsuperscript{1} \and Xin Deng\textsuperscript{2} \and Bing Li\textsuperscript{3} \and Weiming Hu\textsuperscript{3} \and Ali Borji\textsuperscript{4}}


\institute{
            \textsuperscript{1} M. Xu (Corresponding author) and M.L. Qiao are with the School of Electronic and Information Engineering, Beihang University, Beijing, 100191, China (e-mail:MaiXu@buaa.edu.cn;minglangqiao@buaa.edu.cn). \at
            \textsuperscript{2} X. Deng is with the School of Cyber Science and Technology,  Beihang University, Beijing, 100191, China (e-mail:cindydeng@buaa.edu.cn). \at
            \textsuperscript{3} Y.F. Liu, B. Li, W.M. Hu are with National Laboratory of Pattern Recognition, Institution of Automation, Chinese Academy of Sciences; School of Artificial Intelligence, University of Chinese Academy of Sciences and CAS Center for Excellence in Brain Science and Intelligence Technology. \at
            \textsuperscript{4} Ali Borji is with Primer.AI Inc., San Francisco, CA (e-mail: aliborji@gmail.com). \at
            M.L. Qiao and Y.F. Liu contributed equally to this research.
}

\date{}

\maketitle

%

\begin{abstract}
\xma{Visual and audio events simultaneously occur and both attract attention. However, most existing saliency prediction works ignore the influence of audio and only consider vision modality. In this paper, we propose a multi-task learning method for visual-audio saliency prediction and sound source localization on multi-face video by leveraging visual, audio and face information. Specifically, we first introduce a large-scale database of multi-face video in visual-audio condition (MVVA), containing eye-tracking data and sound source annotations. Using this database, we find that sound influences human attention, and conversly attention offers a cue to determine sound source on multi-face video. Guided by these findings, a visual-audio multi-task network (VAM-Net) is introduced to predict saliency and locate sound source. VAM-Net consists of three branches corresponding to visual, audio and face modalities. Visual branch has a two-stream architecture to capture spatial and temporal information. Face and audio branches encode audio signals and faces, respectively. Finally, a spatio-temporal multi-modal graph (STMG) is constructed to model the interaction among multiple faces. With joint optimization of these branches, the intrinsic correlation of the tasks of saliency prediction and sound source localization is utilized and their performance is boosted by each other. Experiments show that the proposed method outperforms 12 state-of-the-art saliency prediction methods, and achieves competitive results in sound source localization.} 
\end{abstract}

\section{Introduction}
\label{intro}

\begin{figure*}[t]
	\begin{center}
		\vspace{-.5em}
		\includegraphics[width=1.0\linewidth]{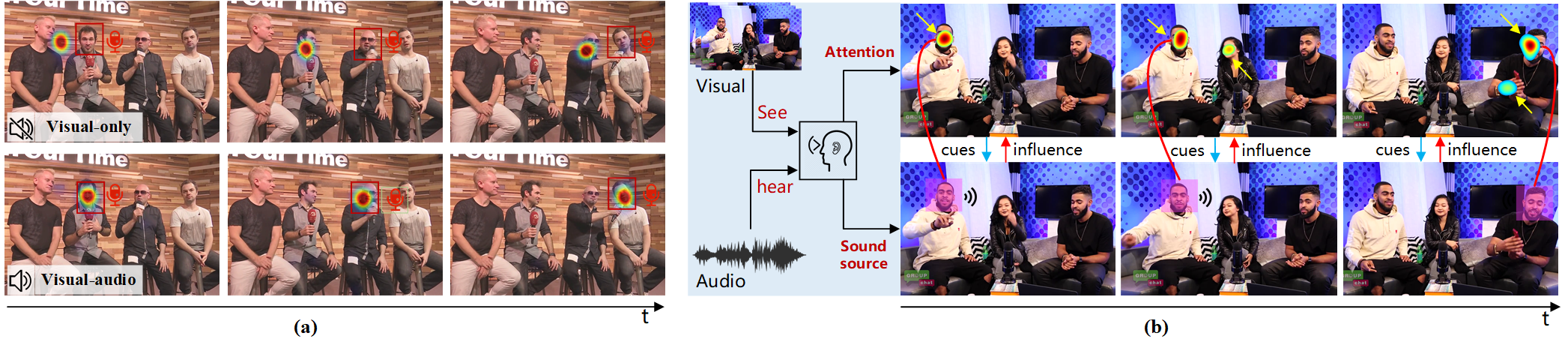}
	\end{center}
	\vspace{-2.0em}
	\caption{(a) An example of visual attention on a multi-face video. Four persons are speaking in a sequence from the left to the right. The first row (``visual-only'') represents the condition when subjects view only mute frames. The second row (``visual-audio'') shows the condition when both visual and audio information is present. (b) An example of the correlation between human attention and sound source localization. \xma{The pink bounding box represents the sound source region.}}
	\label{fig:intro}
	\vspace{-1.0em}
\end{figure*}

\xma{With the rapid development of video platforms, such as YouTube and NetFlix, millions of videos have emerged during the past years. A large proportion of those videos, including movies, video conferences, interviews and variety shows, contain more than one face.
	In multi-face videos, faces are dominate salient objects that attract human attention.
	Therefore, it is important and interesting to model human attention on multi-face videos through saliency prediction. 
	Saliency prediction on multi-face videos has many applications such as video analytics, human-computer interface design, event understanding, perceptual video coding \citep{xu2018find}, \etc
}
\xma{
	During the past few years, the flourish of deep learning has significantly boosted the performance of saliency prediction \citep{wang2018revisiting,jiang2018deepvs,cornia2018predicting,drostejiao2020,huang2015salicon,pan2017salgan,wang2017deep,min2019tased,bak2017spatio}, in particular in multi-face video saliency prediction \citep{liu2017predicting, xu2018find}.
}
\xma{However, deep saliency models only concentrate on visual information, and often ignore auditory information.
	In practice, however, videos are always played with sound, which is an important cue in guiding human attention.
	As illustrated in Fig. \ref{fig:intro} (a), humans pay attention to different regions in presence or absence of sound in the video.
	They fixate at the salient face and transit to other faces faster when sound is available.
	Without sound, people often rely on visual cues (\eg, motion) to locate the speaking person, leading to slower attention transition.
	Therefore, only considering visual information is not sufficient to predict where people look in real-world scenes. }
\xma{
	More importantly, sound source is highly correlated with human attention in multi-face videos. Fig. \ref{fig:intro} (b) presents an example where sound source influences attention, and in return, the attention regions provide cues to localize sound source. 
	Hence, the combination of sound source localization and saliency prediction has potential in improving the performance for both tasks, which has not been considered in previous works.
	Unfortunately, there is little cross talk between existing methods of multi-face video saliency prediction \citep{liu2017predicting, xu2018find} and sound source localization \citep{senocak2018learning,owens2018audio,arandjelovic2018objects,zhao2018sound}. 
}

\xma{Here, we propose a multi-task learning method for visual-aduio saliency prediction and sound source localization on multi-face video, which jointly leverages the information of visual, audio and face. Specifically,  we first establish a large-scale database of multi-face videos in visual-audio condition (MVVA), which includes fixations of 34 subjects and annotated location of sound source on 300 multi-face videos. }
\xma{Then, we mine our MVVA database to obtain several findings. 
	In particular, we find that human attention consistently focuses on one among multiple faces in a video, and that the attention on and its transition across faces are influenced by both visual and audio information. In addition, we also find that human attention can be used to guide the localization of sound source on multi-face videos.}

\xma{Inspired by the above findings, we propose a visual-audio multi-task network (VAM-Net) to predict fixations and locate sound source on multi-face videos. VAM-Net consists of three branches with corresponding modalities: visual, audio and face.} 
\xma{
The input audio is fed to the audio branch for learning the sound-related features.
Subsequently, faces are extracted from the input video using a face detector \citep{zhang2016joint} and then fed to the face branch. In the face branch, both extracted faces and sound-related features from the audio branch are encoded to explore and relate the interaction among multiple faces through a spatio-temporal multi-modal graph (STMG). In our VAM-Net, STMG takes faces, audio and global visual features as nodes, yielding a sound source map for each video frame. 
\xma{STMG is able to accurately predict the sound source locations, by leveraging the powerful capability of graph neural network (GNN) in modeling the relationship between nodes. }
Also, the attention weights, corresponding to each extracted face, are generated and fed to the visual branch. Given the attention weights, the visual branch constructs a two-stream architecture to learn spatio-temporal features for visual saliency prediction on multi-face video. Finally, extensive experimental results show the superiority of the proposed method over state-the-art methods in the main task of saliency prediction and the auxiliary task of sound source localization for multi-face video.
}
\xma{The MVVA database and codes of our method are available at: https://github.com/MinglangQiao/MVVA-Database.}

To the best of our knowledge, this paper is a first attempt to build a multi-task learning framework for saliency prediction and sound source localization. Our main contributions in this paper are three-fold:
\begin{itemize}
	\item \xma{We establish the MVVA database, as a large-scale multi-face video database for visual-audio saliency prediction and sound source localization}.
	\item We thoroughly analyze our MVVA database, study the influence of face and sound on human attention, and explore the factors that impact sound source localization. 
	\item \xma{We propose a deep learning model called VAM-Net that fuses visual, face and audio information to jointly learn the tasks of visual-audio saliency prediction and sound source localization on multi-face video.}
\end{itemize}

\xma{This paper significantly extends our conference paper \citep{ml_salient_face} by jointly learning the tasks of saliency prediction and sound source localization, rather than the single task of saliency prediction \citep{ml_salient_face}. Accordingly, the extension is in the following aspects. 
	1) We supplement a profound analysis on the factors that influence sound source localization, motivating us to embed sound source localization as an auxiliary task for saliency prediction. 
	2) To simultaneously predict saliency and locate sound source, we re-design the multi-task deep learning architecture of VAM-Net, instead of the single task architecture \citep{ml_salient_face}. In addition, STMG, a novel GNN, is added to the VAM-Net to fuse the multi-modal information and explore the interaction among faces.
	Consequently, VAM-Net obtains higher saliency prediction performance than \citep{ml_salient_face}, in particular up to 0.577 gain in normalized scanpath saliency (NSS). It also achieves competitive results on sound source localization.
	3) We conduct additional experiments on both sound source localization and saliency prediction, \eg, comparing with more methods, and evaluating on more databases, as well as additional ablation studies.
}

\section{Related work}\label{sec:related_work}

\vspace{-3em}
\xma{\subsection{Saliency prediction}
	\noindent \textbf{Visual saliency prediction.}} Visual saliency models have been widely developed to predict where people look in images ~\citep{huang2015salicon,zhang2016exploiting,pan2017salgan,wang2017deep,cornia2018predicting,li2014visual} or videos \citep{hossein2015many,bak2017spatio,liu2017predicting,jiang2021deepvs2,wang2018revisiting, min2019tased, zanca2019gravitational,bellitto2021hierarchical, li2010probabilistic,souly2016visual}. 
\xma{
	The seminal work of \cite{itti1998model} proposed a computational model to predict the image saliency, via combining three low-level features including color, intensity, and orientation. 
	\xma{Since then, a number of low-level feature-based saliency prediction methods have emerged \citep{harel2007graph,le2007predicting,xu2016bottom,hossein2015many}.}
	For example, \cite{le2007predicting} proposed to incorporate both the achromatic and chromatic visual features to compute spatial saliency.
	\xma{\cite{harel2007graph} introduced a graph based model leveraging several low-level image features for saliency prediction.}
	\xma{Later, some authors proposed to combine both high- and low-level features to predict human attention \citep{cerf2008predicting,judd2009learning}.}
	For example, \cite{cerf2008predicting} adopt both low-level feature maps (\ie, color, intensity, orientation) and face conspicuity maps to predict human fixations. 
	\xma{\cite{judd2009learning} proposed an SVM method for saliency prediction, which is based on the extracted low-, middle- and high-level image features.}
}

Recently, \xma{visual saliency prediction has achieved a great success, benefiting from the powerful deep neural networks (DNNs) and large scale eye-tracking databases \citep{wang2018revisiting, jiang2018deepvs}
}.
\xma{ In particular, a great number of deep saliency methods \citep{huang2015salicon,wang2017deep,pan2017salgan} use convolutional features to predict visual saliency.
}
\xma{\cite{cornia2018predicting} utilized a dilated convolutional network and an attentive convolutional long short-term memory (LSTM) \citep{xingjian2015convolutional} to extract more sufficient and accurate visual saliency information.
}
\xma{\cite{pan2017salgan} introduced a model based on generative adversarial networks (GAN) to predict saliency.
}
Over videos, most works \citep{wang2018revisiting,liu2017predicting,bak2017spatio,jiang2021deepvs2} integrate CNNs and LSTMs to learn spatial and temporal visual features. 
\xma{\cite{bak2017spatio} proposed a two-stream CNN architecture, the inputs of which are the RGB frames and optical flow sequences, respectively.}
\cite{zanca2019gravitational} leveraged various visual features, such as face and motion, to predict the scanpaths of fixations on images and videos.
Recently, some works have focused on predicting saliency over multi-face videos. \cite{liu2017predicting} proposed an architecture to combine a CNN and a multiple-stream LSTM to learn face features.
\xma{A comprehensive overview of saliency prediction can be found in \citep{borji2012state, borji2018saliency}.}
\xma{Unfortunately}, all of the above methods have discarded the audio modality. In contrast, our method utilizes both audio and video modalities for saliency prediction.

\noindent \xma{\textbf{Visual-audio saliency prediction.}} 
Only a few methods take into account the auditory modality.
The early models \citep{coutrot2014audiovisual, coutrot2015efficient, tsiami2016towards} mainly depend on hand-crafted features. 
In \citep{coutrot2014audiovisual, coutrot2015efficient}, low-level features (\eg, luminance information) and faces are used as visual information, while the audio is fed into a speaker diarization algorithm  to locate the speaking person. Then, the saliency maps of a multi-face video are generated by integrating the visual and audio information. 
\cite{tsiami2016towards} proposed to combine a visual saliency model \citep{itti1998model} and an audio saliency model \citep{kayser2005mechanisms}. However, \citep{tsiami2016towards} only considers the scenario in which a simple stimuli is moving in clustered images.
Recent works tend to make use of machine learning methods.
\xma{\cite{boccignone2018give} proposed a probabilistic framework to predict the saliency maps of conversational scenes, via sampling the attractive locations based on a list of pre-computed priority feature maps. However, this method only considers the simple audio scenes, and it relies on several existing deep learning models \citep{chung2016out,kumar2007profile} to obtain the required features.
}

\xma{For visual-audio saliency prediction, few DNN models have been proposed.
	\xma{
		\cite{jain2020vinet} proposed a 3D convolutional encoder-decoder architecture, named AViNet, to predict visual saliency. In AViNet, SoundNet \citep{aytar2016soundnet} is applied to extract audio features and S3D \citep{xie2018rethinking} for visual features, which are fused to output saliency maps of videos. 
	}
	Most recently, \cite{tavakoli2019dave} have developed a two-stream 3D-CNN {\citep{hara2018can}} to encode visual and audio information into feature vectors, which are then concatenated to learn visual-audio saliency.
	\cite{tavakoli2019dave} do not focus on saliency prediction of multiple faces in a video, which is the main target of our work.
	More importantly, we develop a brand-new multi-task DNN architecture for jointly learning to predict saliency and locate sound source in multi-face videos.
}

\subsection{\xma{Sound Source localization}}
\xma{
	Sound source localization in visual context aims at locating the spatial regions that make sound in images and videos.
	Recently, several deep learning methods \citep{pami_avm_8894565,senocak2018learning,owens2018audio,arandjelovic2018objects,zhao2018sound,tian2018audio,hu2020discriminative,jia2020look} have been proposed for sound source localization, achieving remarkable progress.
	Among them, several methods utilize the synchrony or correspondence of visual and audio signals to train the DNN models, and then employ visualization algorithms to obtain the sound source heat map from the DNN models. 
	Note that visualization can be conducted by various algorithms, including directly showing the feature maps, computing maps with class activation map (CAM) \citep{zhou2016learning}, and other similar algorithms.
	For example, \cite{owens2018audio} proposed a self-supervised algorithm that trains a DNN to predict whether the video frames and audio waves are aligned in the temporal domain. 
	Then, the sound source map is obtained by applying the CAM visualization algorithm.
	In \citep{arandjelovic2018objects}, an audio-visual correspondence network was designed to localize objects that make sound in images.
	\cite{zhao2018sound} introduced a cross-modal learning system, named PixelPlayer, to achieve the localization and separation of sounds.
	\cite{pami_avm_8894565} used an attention mechanism to explore the correlation between visual and audio modalities.
	They developed a two-stream architecture consisting of a visual subnet and an audio subnet. Then, a localization subnet is leveraged to integrate the two-stream features and to locate sound source regions in images.
}

\xma{The VAM-Net, as a multi-task learning method, is proposed for simultaneously predict visual-audio saliency and to locate sound sources. 
	Our VAM-Net method differs from the traditional sound source localization methods in two aspects: 1) VAM-Net guides the localization of sound source by making use of human attention, and 2) A novel GNN, \ie, STMG, is proposed in the VAM-Net to fuse the multi-modal information of face and sound, for locating sound in multi-face video. 
}

\subsection{Visual-audio databases}

\xma{
	Only few databases are available for studying visual-audio attention \citep{coutrot2013toward, coutrot2014saliency,coutrot2015efficient}. 
	\xma{
		The details about these databases are summarized in Tab. 1 of the supplemental material.
	}
	These datasets are limited in the following ways.
	First, they are small. 
	In particular, the numbers of videos in these databases are typically under 150, which is insufficient to train DNNs. 
	Second, their videos 
	contain only one or a few scenes. 
	For example, Coutrot II \citep{coutrot2014saliency} and Coutrot III \citep{coutrot2015efficient} only include conversation and 4 person meetings events. 
	Third, all of their videos have low resolution.
	To be specific, their resolutions are up to $1232 \times 504$, lower than the high definition standard (\ie,  $1920 \times 1080$ or $1280 \times 720$).
	More importantly, to the best of our knowledge, none of the visual-audio eye-tracking databases contain both eye-tracking data and annotated sound source.
}

\xma{
	For sound source localization, the existing databases are diverse in annotation style, content and scale. On the one hand, several databases \citep{arandjelovic2018objects, pami_avm_8894565,hu2020discriminative,jia2020look} have been proposed mainly for instrument scenes, annotated in the form of image-audio pair, \ie, one labeled frame and the corresponding audio clip. For example, \cite{jia2020look} introduced a sound source database called INSTRUMENT-32CLASS, which contains 3,604 image-audio pairs with 32 instrument classes and only 747 pairs are annotated by a segment mask. \cite{hu2020discriminative} collected 3 larger sound localization databases comprised of more than 29,000 image-audio pairs over 15 instrument classes. The labeled images are annotated by bounding boxes. 
	On the other hand, a few databases concentrate on multi-face videos \citep{chakravarty2016cross,roth2020ava}. They usually annotate bounding boxes on faces with speaking/non-speaking labels.
	For instance, \cite{chakravarty2016cross} built a 35-minute multi-face video database in a specific scene, in which the videos are all cut from a panel discussion video.
	Subsequently, a larger database \citep{roth2020ava} containing 160 videos with about 40,000 labeled face tracks was established. It is currently the largest active speaker detection database over multi-face videos.}

\xma{Different from the above databases, our MVVA database contains both eye-tracking and sound source labels of multi-face videos with diverse scenes.
	Furthermore, MVVA has annotated each frame from all 300 videos, which have a total number of 146,000+ frames and 923 labeled face tracks.
	Our database is publicly available online to facilitate the future research on visual-audio saliency prediction and sound source localization.
}

\begin{figure*}[htbp]
	\centering
	\includegraphics[width=1.0\linewidth]{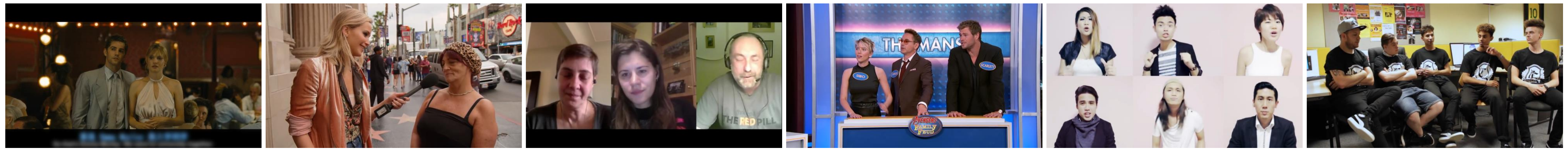}
	\vspace{-1.5em}
	\caption{An example frame from each category of videos considered here. From left to right, the videos belong to TV play/movie, interview, video conference, TV show, music/talk show, and group discussion.}
	\label{fig:Video_class_example}
	\vspace{-1em}
\end{figure*}

\section{The Proposed Database}
In this section, we introduce a large-scale eye-tracking database called multi-face video in visual-audio condition (MVVA). The proposed database contains eye-tracking fixations when both audio and video were presented. \mla{Besides, we have manually annotated all talking faces at the frame level for all videos.} To the best of our knowledge, \xma{our database is the first public database that has multi-face videos with audio information and contains both eye-tracking data and sound source annotations.}
\mla{Therefore}, in addition to saliency, it can be used in other research areas such as sound localization, \mla{speaker diarization}, \emph{etc}, since the faces of speakers are manually marked.
\xma{Here, we present the details about the database creation as follows.}

\noindent\textbf{Stimuli.} A total number of 300 videos with 146,529 frames, containing both images and audio, were collected. Among them, 143 videos were selected from MUFVET \citep{liu2017predicting} and other 157 videos were selected from YouTube. 
\xma{The selection criterion are as follows.}
\begin{itemize}
	\item \xma{Containing at least one obvious face. 
		Face is an important factor that attracts human attention. We aim to predict the salient face and saliency transition among faces. Thus, we collect videos containing at least one obvious face. In MVVA, the average face number is 3, with the average face size of 101 $\times$ 145 pixels, ranging from 18 $\times$ 20 to 330 $\times$ 457 pixels. }
	
	\item \xma{Diverse audio scenes. 
		As Tab.~\ref{tab:audio_video_Class} shows, our database contains 6 types of audio scenes
		(\ie, the laughter, music, crowd, street, applause and noise).
		It provides abundant data for investigating the correlation among audio, video, and human attention.
		Besides, we selected audios with Chinese and English languages, to guarantee that the subjects can understand the audio information. 
		Specifically, there are 116 Chinese videos, 179 English videos, and 5 other language videos.
	}
	
	\item \xma{High video quality. Selected videos have a resolution of 1280 $\times$ 720, with the frame rate ranging from 24 to 30 fps (27 fps on average).  All videos were carefully checked to ensure high visual quality, and have a total length of 5,357 seconds. Some examples of the selected videos are shown in Fig. \ref{fig:Video_class_example}. }
	
	\item \xma{Diverse visual scenes. To ensure scene diversity, the selected videos belong to 6 main categories: TV play/movie, interview, talk show, video conference, music/talk show and Group discussion. The scene diversity is rather important for a database to evaluate the generalization performance of different models. The detailed statics of the scenes in our MVVA database can be found from Tab. \ref{tab:audio_video_Class}.}
\end{itemize}

All of the videos were encoded by H.264 with duration varying from 5 to 30 seconds. 
Note that these 300 videos are either indoor or outdoor scenes, and can be classified into 6 categories as mentioned above. 
The audio content covers different scenarios including quiet scenes (\eg, news broadcasting) and noisy scenes (\eg, interview in subway and talking in a party).

\begin{table}[htbp]
	\centering
	\caption{Video categories and audio scenes in our database.}
	\vspace{-1em}
	\setlength{\tabcolsep}{0.1mm}{%
		\scriptsize%
		\begin{tabular}{c|cccccc|c} 
			\hline
			\multicolumn{1}{p{4.6em}|}{\makecell*[c]{\bfseries Video\\\bfseries category}} & \multicolumn{1}{p{4.5em}|}{\makecell*[c]{TV paly/\\Movie}} & \multicolumn{1}{p{4.1em}|}{\makecell*[c]{\makecell*[c]{Interview}}} & \multicolumn{1}{p{5.em}|}{\makecell*[c]{Video\\conference}} & \multicolumn{1}{p{4.6em}|}{\makecell*[c]{TV show}} & \multicolumn{1}{p{4.5em}|}{\makecell*[c]{Music/\\Talk show}} & \multicolumn{2}{p{6.2em}}{\makecell*[c]{Group\\discussion}} \\
			\hline
			\multicolumn{1}{p{4.6em}|}{\makecell*[c]{\bfseries Number}} &   \multicolumn{1}{p{4.5em}|}{\makecell*[c]{53}}    &   \multicolumn{1}{p{4.1em}|}{\makecell*[c]{71}}   &    \multicolumn{1}{p{5.em}|}{\makecell*[c]{14}}   &    \multicolumn{1}{p{4.8em}|}{\makecell*[c]{67}}   &   \multicolumn{1}{p{3.955em}|}{\makecell*[c]{51}}    &   \multicolumn{2}{p{4.89em}}{\makecell*[c]{ 11}} \\
			\hline
			\hline
			\multirow{2}{*}{\makecell*[c]{\bfseries Audio\\\bfseries scenes}} & \multicolumn{6}{p{22.67em}|}{\makecell*[c]{Noisy scenes}}    & \multirow{2}{*}{\makecell*[c]{Quiet\\scenes}} \\
			\cline{2-7}    \multicolumn{1}{c|}{} & \multicolumn{1}{c|}{Laughter} & \multicolumn{1}{p{4.1em}|}{\makecell*[c]{Street}} & \multicolumn{1}{p{5.0em}|}{\makecell*[c]{Music}} & \multicolumn{1}{p{4.6em}|}{\makecell*[c]{Applause}} & \multicolumn{1}{p{4.5em}|}{\makecell*[c]{Crowd}} & \multicolumn{1}{p{2.8em}|}{\makecell*[c]{Noise}} & \\
			\hline
			\multicolumn{1}{p{4.6em}|}{\makecell*[c]{\bfseries Number}} &   \multicolumn{1}{c|}{34}    &   \multicolumn{1}{p{4.1em}|}{\makecell*[c]{17}}    &   \multicolumn{1}{p{5.0em}|}{\makecell*[c]{72}}    &   \multicolumn{1}{p{4.8em}|}{\makecell*[c]{16}}    &   \multicolumn{1}{p{3.955em}|}{\makecell*[c]{46}}    &   \multicolumn{1}{p{2.8em}|}{\makecell*[c]{19}}    &  \multicolumn{1}{c}{\makecell*[c]{96}}  \\
			\hline
		\end{tabular}%
	}
	\label{tab:audio_video_Class}
\end{table}%

\noindent\textbf{Apparatus.} For monitoring the binocular eye movements the EyeLink 1000 Plus \citep{eyeLink1000} eye tracker was used in our experiment. EyeLink1000 Plus is an integrated eye tracker with a 23.8'' TFT monitor at screen resolution of 1280 $\times$ 720. During the experiment, EyeLink1000 Plus captured gaze data at 500 Hz.
\xma{We used the pupil-corneal reflection (Pupil-CR) tracking mode to ensure the robustness and high accuracy of eye-tracking. During experiments, the eye tracker worked in remote mode, in which the subjects can view the screen freely without fixating their heads on a tower mount. }
According to \citep{eyeLink1000}, the gaze accuracy can reach 0.25-0.5 visual degree in the head \xma{remote mode}. \xma{The visual content and audio signal were
	synchronized during experiments, with support of the experiment builder software accompanied with the eye tracker. We manually checked all videos to ensure that there is no perceptual latency when playing video and sound. During experiments, the audio was played using an earphone, and the volume can be adjusted by the subjects for clear hearing. See \citep{eyeLink1000} for more details about EyeLink1000 Plus.}

\begin{figure}[t]
	\begin{center}
		\vspace{-.8em}
		\includegraphics[width=1.0\linewidth]{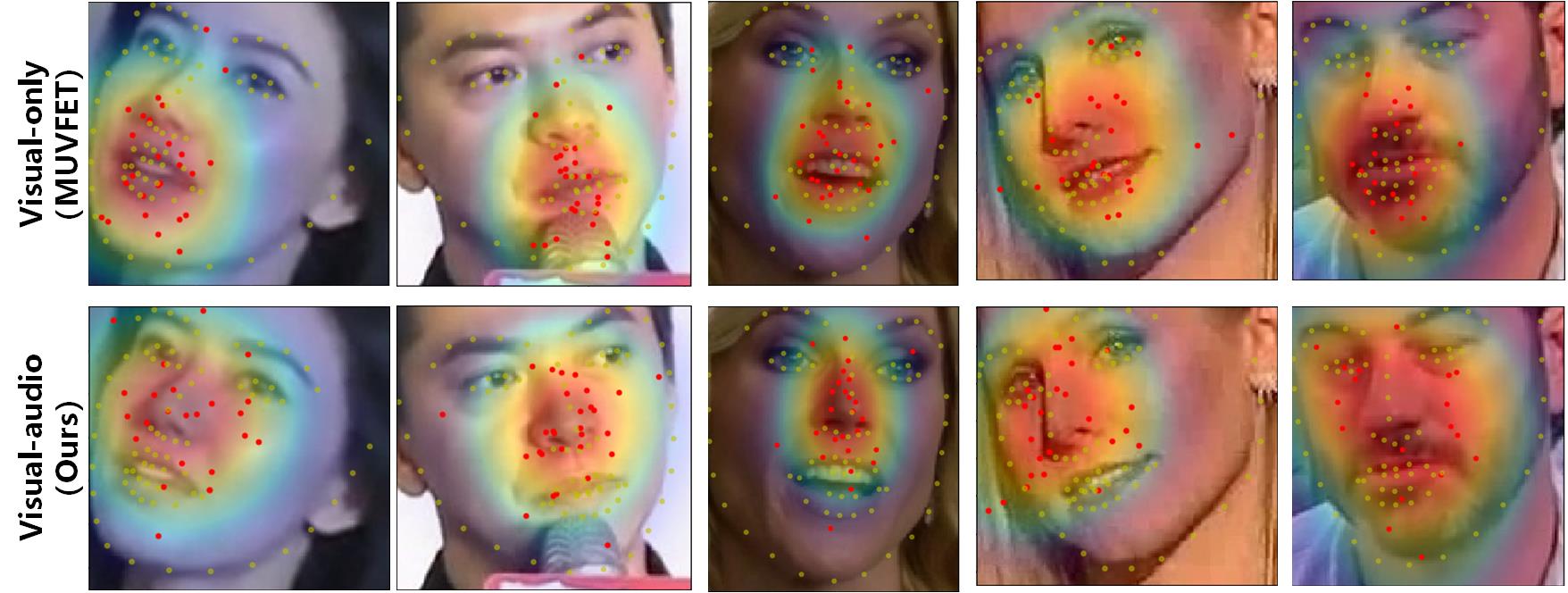}
	\end{center}
	\vspace{-1.0em}
	\caption{Examples of saliency maps in visual-only (the first row) and visual-audio condition (the second row). Note that the red dots are fixation points, and the light yellow dots are facial landmarks.}
	\label{fig:attention_example}
	\vspace{-1.5em}
\end{figure}

\noindent\textbf{Participants.} 34 participants (21 males and 13 females), aging from 20 to 54 (24 in average), were recruited to participate in the eye-tracking experiment.
All participants had normal \mla{or} corrected-to-normal vision. \xma{Among the participants, 32 are naive viewers without any knowledge about the eye-tracking experiments, and 2 had prior experience with similar experiments.} It is worth pointing out that only subjects who passed the eye tracking calibration were quantified for the experiment. Finally, 34 subjects (out of 39) were selected to participate in our experiment. \xma{This number is sufficient for eye-tracking experiments, according to the conclusion of \citep{jiang2021deepvs2}. }

\noindent\textbf{Procedure.} During eye tracking, subjects were required to sit on a comfortable chair with the viewing distance of $\sim 55cm$ from the screen. Before viewing the videos, each subject was required to perform a 9-point calibration for the eye tracker. \xma{Next, a validation procedure was performed for initial calibration, and to ensured that the subject is able to re-fixate the targets \citep{eyeLink1000}. If the subject did not pass the validation procedure, a re-calibration was required; otherwise, it continued to the next step}. \xma{After the validation}, videos were shown in a random order and subjects were asked to view them freely.
Besides, a 5-second blank period with a black screen was inserted between each two successive videos for a short break.
Note that the audio and video stimuli were presented simultaneously during the experiment. In order to avoid eye fatigue, the 300 videos were equally divided into 6  \xma{equal} sessions \xma{with similar content}, and there was a 5-minute rest after viewing each session. \xma{Before each session, the calibration and validation procedures were performed as aforementioned}. 
\xma{The entire experiment last about 2.5 hours for each subject.}
In total we collected 5,013,980 fixations over all 34 subjects and 300 videos.

\noindent\textbf{\xma{Talking-face annotation.}} \xma{In order to investigate the correlation between human attention and the talking face, we 
	annotated the talking face in all videos on each frame. Specifically, we first used a state-of-the-art face detection model \citep{zhang2016joint} to locate the faces of each frame and assigned each face a numeric ID. Then, we recruited 7 subjects (5 undergraduates and 2 postgraduates) to complete the talking-face annotation. In particular, the 5 undergraduates were asked to annotate the talking face in each video; then, the annotated results were checked and corrected by two postgraduates. Meanwhile, the types of sounds were also annotated, including speaking, laughter/applause and singing. In addition to saliency prediction, our database can be used for some other tasks, such as the sound localization \citep{owens2018audio,senocak2018learning,arandjelovic2018objects}, multi-modal event detection \citep{tian2018audio} and sound separation \citep{gao2018learning}.
}

\begin{figure}[t]
	\centering
	\vspace{-1.0em}
	\hspace{-1.2em}
	\subfloat[]{
		\begin{minipage}[t]{0.49\linewidth}
			\centering
			\label{fig:faceFeature_nss}
			\includegraphics[width=1.7in]{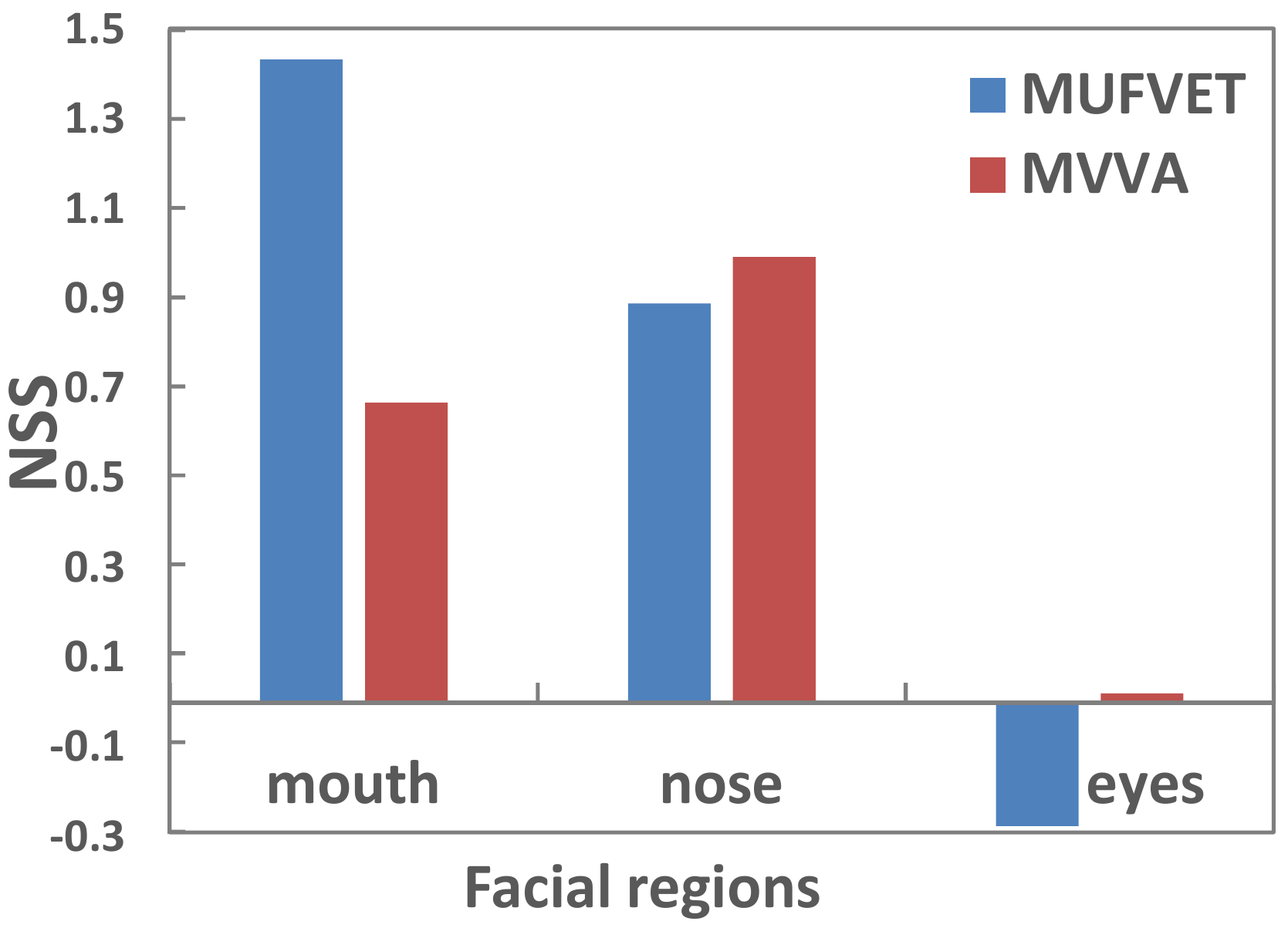}
		\end{minipage}%
	}%
	\subfloat[]{
		\hspace{.2em}
		\begin{minipage}[t]{0.49\linewidth}
			\centering
			\label{fig:contextual_NSS}
			\includegraphics[width=1.7in]{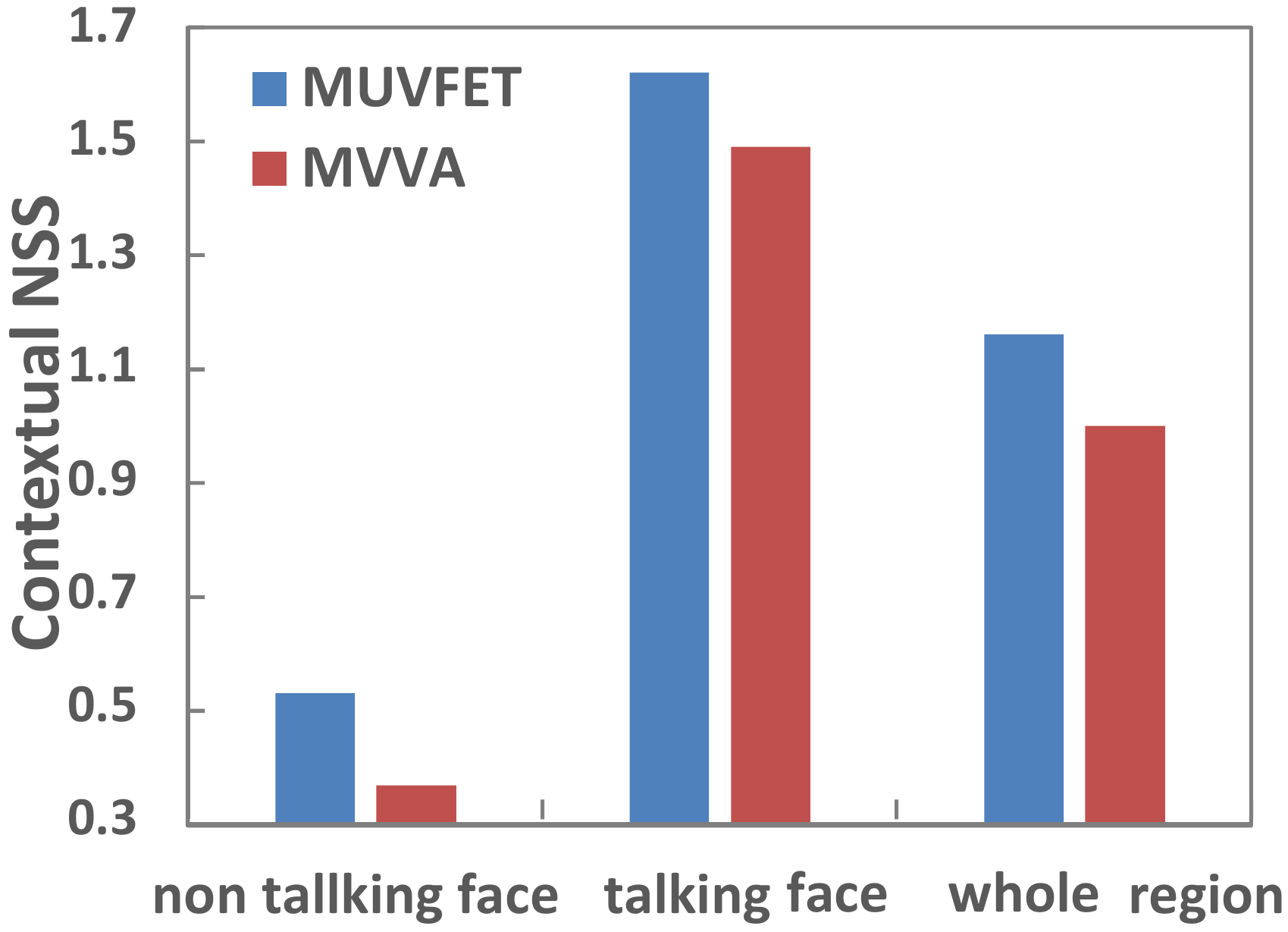}
		\end{minipage}%
	}%
	\centering
	\vspace{-1em}
	\caption{(a) NSS of saliency on different facial landmarks in visual-only (MUVFET)/visual-audio (ours) conditions.
		(b) Contextual NSS of optical flow maps over different face regions.}
	\vspace{-1.5em}
\end{figure}

\section{Database analysis} \label{sect:Database analysis}
\xma{In this section, we mainly focus on analyzing human attention and sound source localization for multi-face videos, when presenting both audio and video.
	Here, our analysis is based on our MVVA database for visual-audio saliency and the MUFVET database for audio-only saliency.
}

\subsection{Consistency analysis of human attention}
\xma{First, we measure the consistency of human attention on multi-face videos, with the following finding. }

\noindent\textit{\xma{Finding 1: The attention of subjects is consistent on multi-face videos, in particular on the same face, when simultaneously presenting audio and video.}}
 
\xma{\textit{Analysis}: We randomly and equally divide the subjects into two non-overlapping groups (A and B) by 20 trails. Then, the linear correlation coefficient (CC) between the fixations of groups A and B is calculated. The averaged CC value is 0.75 with the standard deviation of 0.08 over our MVVA database,
	close to the averaged CC value of 0.80 with the standard deviation of 0.07 over MUFVET database.
	This implies high consistency across subjects in viewing multi-face videos, when simultaneously presenting audio and video.
	The proportion of the fixations falling into the same face is 73.8\% over our database. We conclude that people tend to concentrate on the same face, when simultaneously presented with audio and video. 
}

\begin{figure}[t]
	\begin{center}
		\vspace{-1.em}
		\hspace{-1.2em}
		\includegraphics[width=1.02\linewidth]{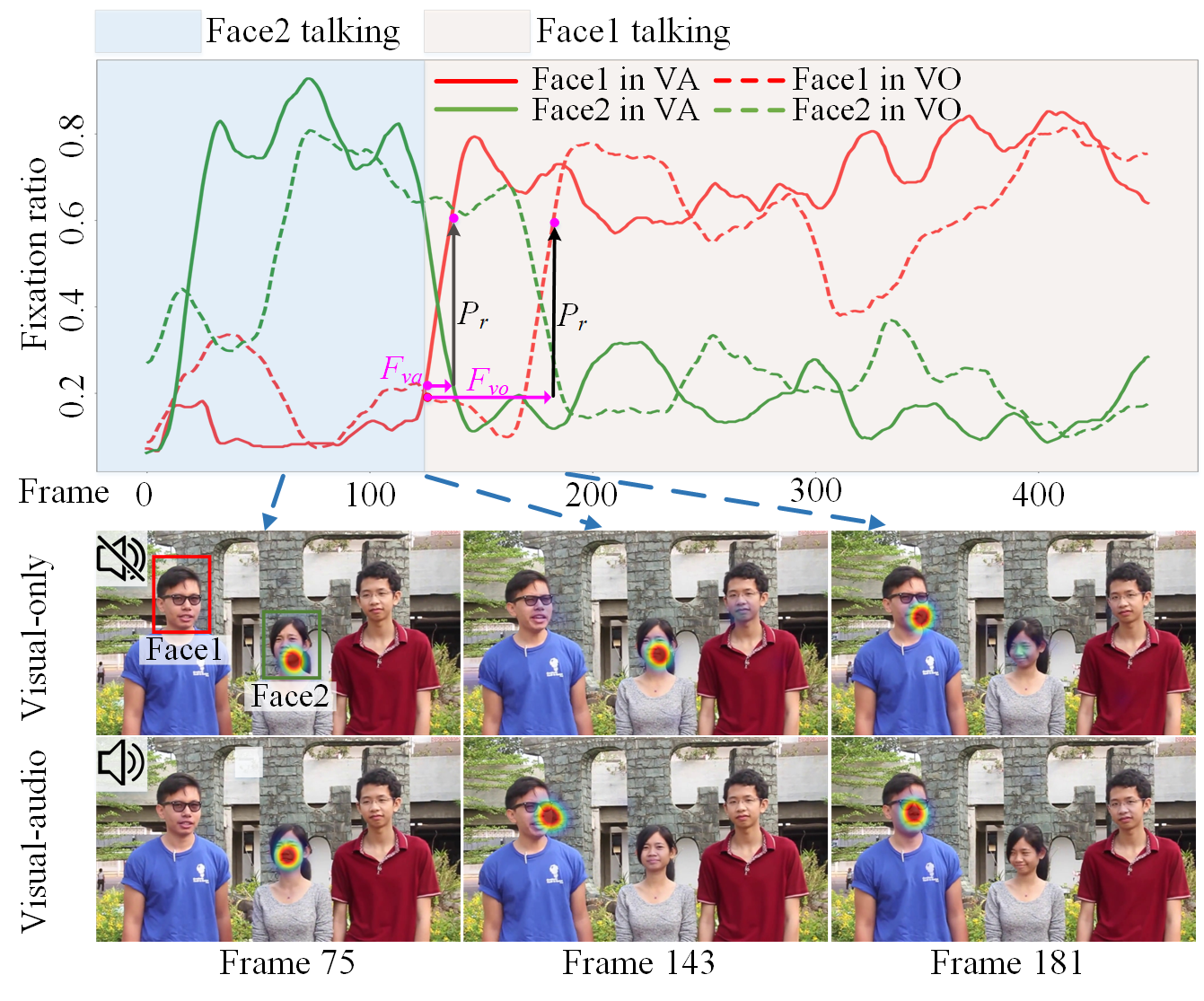} 
	\end{center}
	\vspace{-1.0em}
	\caption{An example of fixation transitions in Visual-Only (VO, the first row of heat maps) and Visual-Audio condition (VA, the second row of heat maps).}
	\label{fig:finding3_vis}
	\vspace{-1em}
\end{figure}

\subsection{The influence of audio on human attention}
\xma{Next, we investigate the influence of audio information on human attention from the aspects of the fixation distribution on faces and the fixation transition across faces.
	We further investigate the influence of motion on attention in the absence of audio.
	We came across the following finding, which inspires the design of our DNN model for the task of visual-audio saliency prediction. } 

\noindent\xma{\textit{Finding 2: In presence of audio, the distribution of fixations on faces is different from that of visual-only scenario.}}

\xma{\textit{Analysis:}
	For quantifying the fixation distribution on faces, we follow \citep{marighetto2017audio} to calculate the averaged dispersion values for the saliency maps of face regions on
	the same videos but at the visual-audio condition (over the MVVA database) and the visual-only condition (over the MUVFET database), respectively.
	The averaged dispersion for the visual-audio condition is 44.06, while that for the visual-only condition is 39.34. 
	This indicates that the fixation distribution on faces at the visual-audio condition is different from that of the visual-only condition.
	We further investigate where fixations distribute on the face region at the visual-audio and visual-only conditions.
	Fig.~\ref{fig:attention_example} shows that \xma{humans tend to fixate at the center of the face (\ie, near nose) in visual-audio condition, while people normally concentrate on the mouth in the visual-only condition.
		To quantify this observation, we follow \citep{tavakoli2019dave} to measure the contextual NSS between the ground truth (GT) saliency maps and the landmarks of mouth, nose and eyes.
		The results of contextual NSS averaged over the same videos of the two databases are shown in Fig. \ref{fig:faceFeature_nss}. 
		We find that our MVVA database has the highest NSS values on nose, while the MUVFET database has the highest NSS values on mouth.
	}
}

\begin{figure}[t]
	\begin{center}
		\includegraphics[width=.98\linewidth]{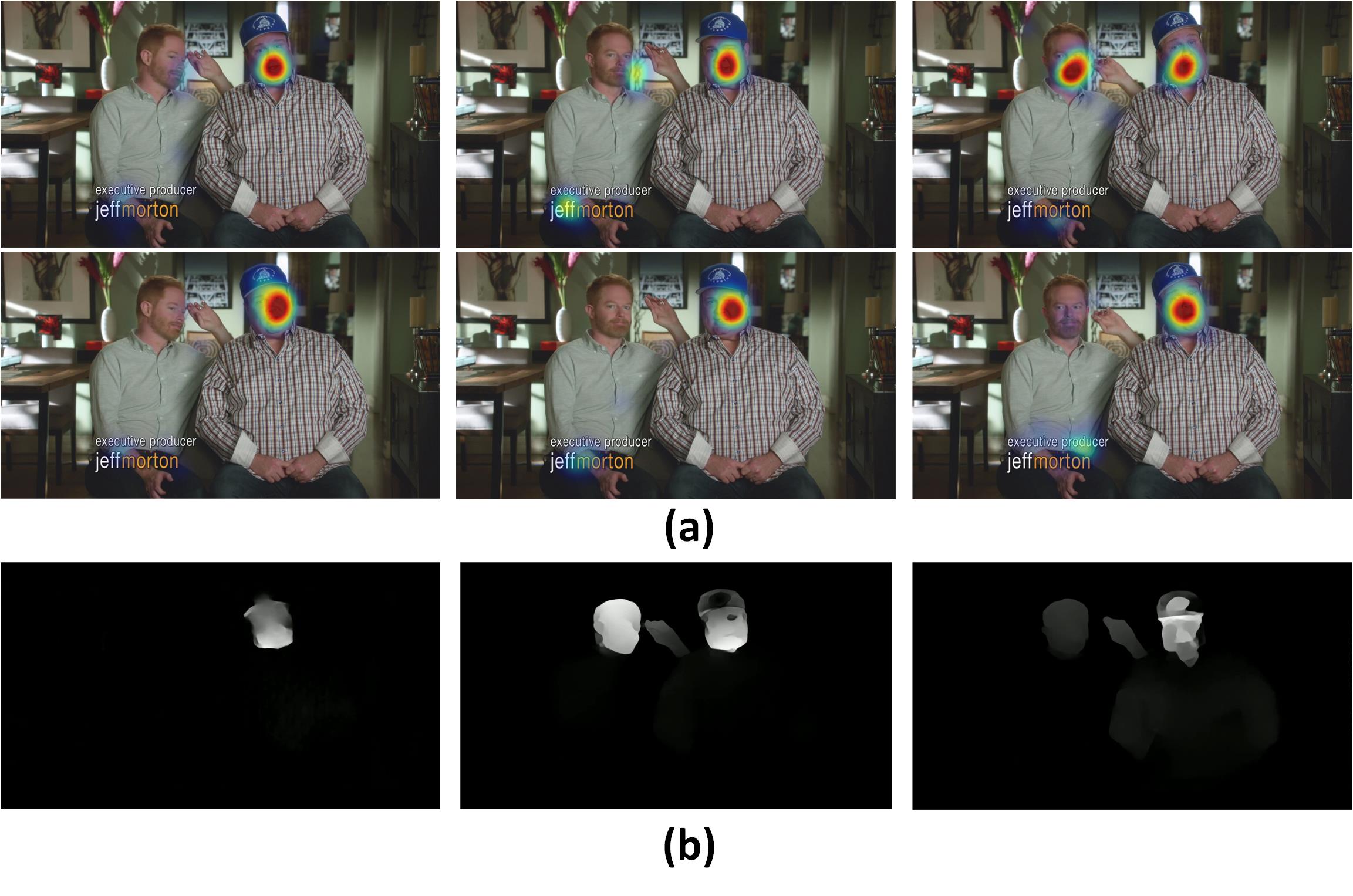}
	\end{center}
	\vspace{-1.5em}
	\caption{(a) An example video showing the saliency difference between Visual-Only condition (1st row) and Visual-Audio condition (2nd row). The person on the right is talking while the other is turning his head. (b) Optical flow map for each frame.}
	\label{fig:face_opticalFlow}
	\vspace{-1em}
\end{figure}

\xma{
	\textit{Finding 3: In the turn-taking scenes, the transition of fixations across faces is largely influenced by audio.}
}

\textit{Analysis:} 
\xma{
	Fig. \ref{fig:finding3_vis} shows an example of attention transition in the turn-taking scenes. It can be observed that human fixations transit and follow the talking face faster in the visual-audio condition than that in the visual-only condition.  Fig.\ref{fig:intro} also shows the similar observation. 
	For quantitative analysis, we compare the attention transition time in visual-audio and visual-only conditions. In particular, we define the attention transition time by the average number of frames that fixations transit to the talking face, when turn-taking happens. 
	Here, $F_{\text{va}}$ and $F_{\text{vo}}$ denote the attention transition time in MVVA (visual-audio condition) and MUVFET (visual-only condition), respectively. The results of $F_{\text{va}}$ and $F_{\text{vo}}$ are 24 and 30 frames, respectively. Thus, the attention transition time in visual-audio condition is shorter than that in visual-only condition by $25\%$. 
	From the above results, we can conclude that the fixations transit across faces are largely influenced by audio.
}

\noindent\textit{Finding 4: Human attention is more influenced by motion in the absence of audio.}

\textit{Analysis:}
\xma{
	Fig. \ref{fig:face_opticalFlow} visualizes the influence of motion on human attention at the visual-only and visual-audio conditions over a sample video.
	We observe that in the absence of audio, 1) attention is mostly attracted by the person on the left hand side who is turning his head, and 2) subjects only concentrate on the speaking person on the right hand side.
	This indicates that people are guided by the visual cue of motion more in the visual-only condition, compared to that in the visual-audio condition.
	We further measure the contextual NSS \citep{tavakoli2019dave} between the heat maps of the magnitude of optical flow and GT fixations in three regions, \ie, the talking face region, the non-talking face region and the whole region. Fig. \ref{fig:contextual_NSS} shows the averaged results of the contextual NSS over the two databases. 
	We can see from this figure that the contextual NSS at the visual-only condition is larger than that at the visual-audio condition over different regions.
	This implies that human attention is more influenced by motion in the absence of audio. Therefore, \textit{Finding 4} is validated.
}

\subsection{Important factors for sound source localization}
\xma{Finally, we investigate the factors for sound source localization over multi-face videos and use our findings to develop a DNN model for sound source localization.}

\noindent\textit{\xma{Finding 5: Fixations are attracted by sound source regions; but they do not always concentrate on sound source regions.}}

	\textit{Analysis:}
	We investigate the correlation between sound source and human attention on multi-face video, by computing the CC between the sound source heat map
	and the fixation map in our database.
	\xma{Specifically, we generate a 2D Gaussian distribution for each talking face at each frame as the sound source map, based on the talking-face annotation.}
	The CC between the sound source map and the fixation map is 0.65, significantly higher than 0. This indicates that human attention is attracted by sound source regions. 
	On the other hand, the consistency between fixations and sound source regions is considerably smaller than 0.75 of human attention consistency mentioned in \textit{Finding 1}.
	This further implies that human fixations do not always concentrate on sound source regions.
	It is probably because there are other visual factors influencing saliency, such as motion and text. 

\begin{figure*}
	\centering
	\vspace{-1.0em}
	\includegraphics[width=1\linewidth]{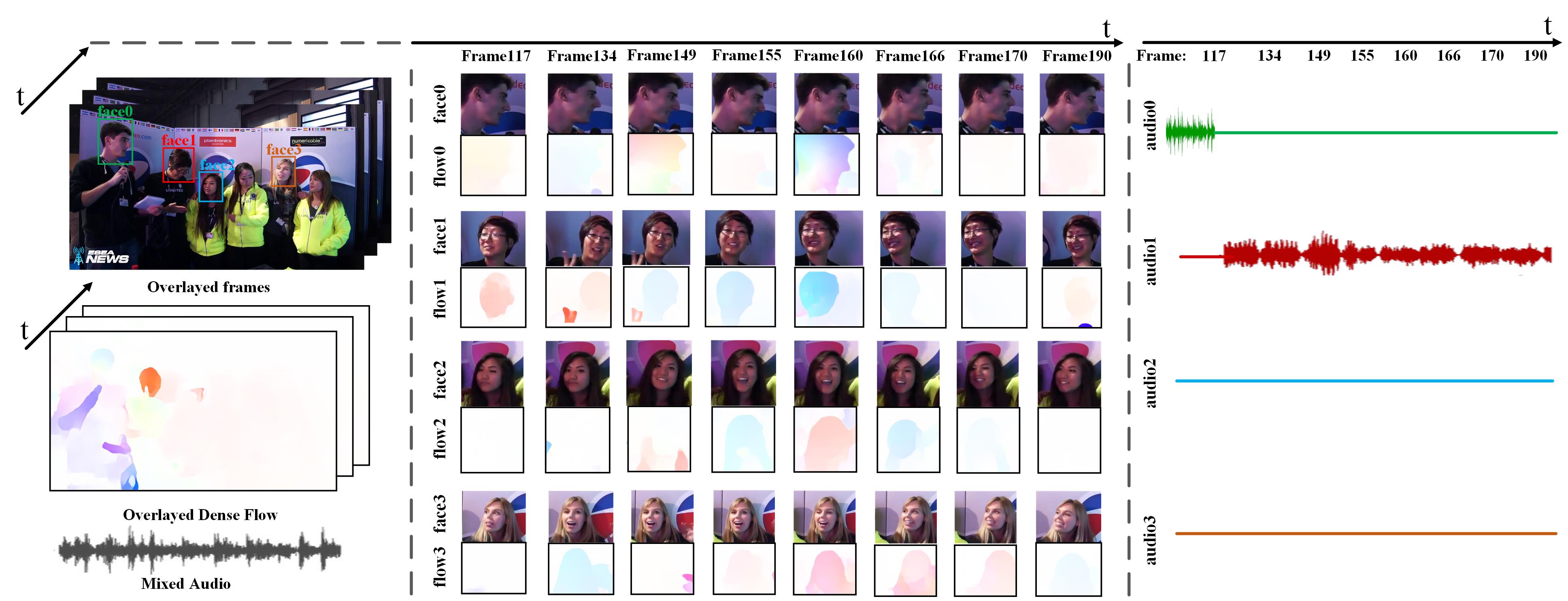}
	\vspace{-1em}
	\caption{\xma{An example of multi-face video conversation scene, in which the visual information is not sufficient for sound source localization. The left part shows the raw video frames and their optical flow maps. The middle part shows the corresponding images and optical flow maps in face regions. The right part illustrates the audio waveform of each face, in which the straight line expresses that the corresponding face is silent.}}
	\label{finding6_conversation}
	\vspace{-1em}
\end{figure*}

\noindent\textit{\xma{Finding 6: Visual information is necessary but not sufficient for sound source localization in multi-face videos.}}

\xma{\textit{Analysis:} 
	Since the speaking faces with mouth motion are the dominant sources of sound, visual information is obviously necessary for sound source localization in multi-face videos.
	More importantly, it is interesting to investigate whether the visual information is sufficient for sound source localization over multi-face videos.
	See Fig. \ref{finding6_conversation} as an example: from frames 134 to 190, faces 1, 2 and 3 have motion in their mouth regions, but only face 1 is the  source of sound. 
	Therefore, visual information is not sufficient for localizing the sound source, and the audio information is also necessary for sound source localization. }
\xma{
	We further quantitatively analyze the effect of audio in sound source localization by comparing the results of sound source localization using visual-audio and visual-only information, respectively. 
	To this end, we take the bounding boxes of manually annotated talking faces as the GT of sound source, denoted by $\mathbf{B}_{\text{gt}}$. Then, we recruited a group of subjects to label the sound source boxes in the visual-only condition (denoted as $\mathbf{B}_{\text{vo}}$) and in the visual-audio condition (denoted as $\mathbf{B}_{\text{va}}$). Here, we compute the Mean Overlap (MO) values between $\mathbf{B}_{\text{gt}}$ and $\mathbf{B}_{\text{vo}}$ over all videos of our MVVA databases:
	\begin{equation}
			\label{mo-defination}
			\text{\text{MO}}_{\text{vo}} =\frac{A(\mathbf{B}_{\text{gt}} \cap \mathbf{B}_{\text{vo}})}{A(\mathbf{B}_{\text{gt}} \cup \mathbf{B}_{\text{vo}})},
	\end{equation}
	where $A(\cdot)$ represents the area of each box.
	Similarly, the MO values $\text{\text{MO}}_{\text{va}}$ between $\mathbf{B}_{\text{gt}}$ and $\mathbf{B}_{\text{va}}$ are calculated over our MVVA database.  
	The results of $\text{\text{MO}}_{\text{vo}}$ and $\text{\text{MO}}_{\text{va}}$ are \mla{0.80} and 0.91, respectively. 
	This gap implies that visual information is not sufficient for sound source localization in multiple face videos. 
}

\xma{From the above findings, we conclude that both visual and audio information are necessary and useful for the tasks of saliency prediction and sound source localization in multi-face videos. 
	Additionally, our findings indicate that the above two tasks share some common characteristics, \eg, both of them are correlated with the talking face; but they also have different emphases, \eg, the task of sound source localization cannot be accomplished by only predicting saliency.
	This suggests that we can apply a multi-task learning framework to simultaneously predict saliency and locate sound source, such that these two tasks can help each other for better performance.
}

\section{The Proposed Method}
\xma{In this section, we present the details about the proposed method, in light of our analysis in Sec. \ref{sect:Database analysis}. We first describe the overall framework in Sec. \ref{framework}. Then, we introduce the architectures of visual, audio and face branches in Sec. \ref{sec:visual_branch}, \ref{sec:audio_branch} and \ref{sec:face_branch}, respectively. Finally, the loss functions and the training protocol are discussed in Sec. \ref{sec:lloss_protocol}.
}

\vspace{-20pt}

\mla{\subsection{Framework}\label{framework}}
According to the above findings, visual information, audio and faces are all important factors that influence human attention.
\xma{It is thus necessary to leverage these multi-modal information for saliency prediction.
	In particular, we find that sound influences human attention, and attention offers a cue to find the sound source in mutli-face videos.
	Therefore, we propose a visual-audio multi-task network (VAM-Net) to simultaneously predict human attention and locate sound source.
	VAM-Net takes video frames, faces and audio signal as input, and outputs saliency maps and sound source heat maps, respectively.
} 

\begin{figure*}[htbp] 
	\centering
	\vspace{-.5em}
	\includegraphics[width=1.0\linewidth]{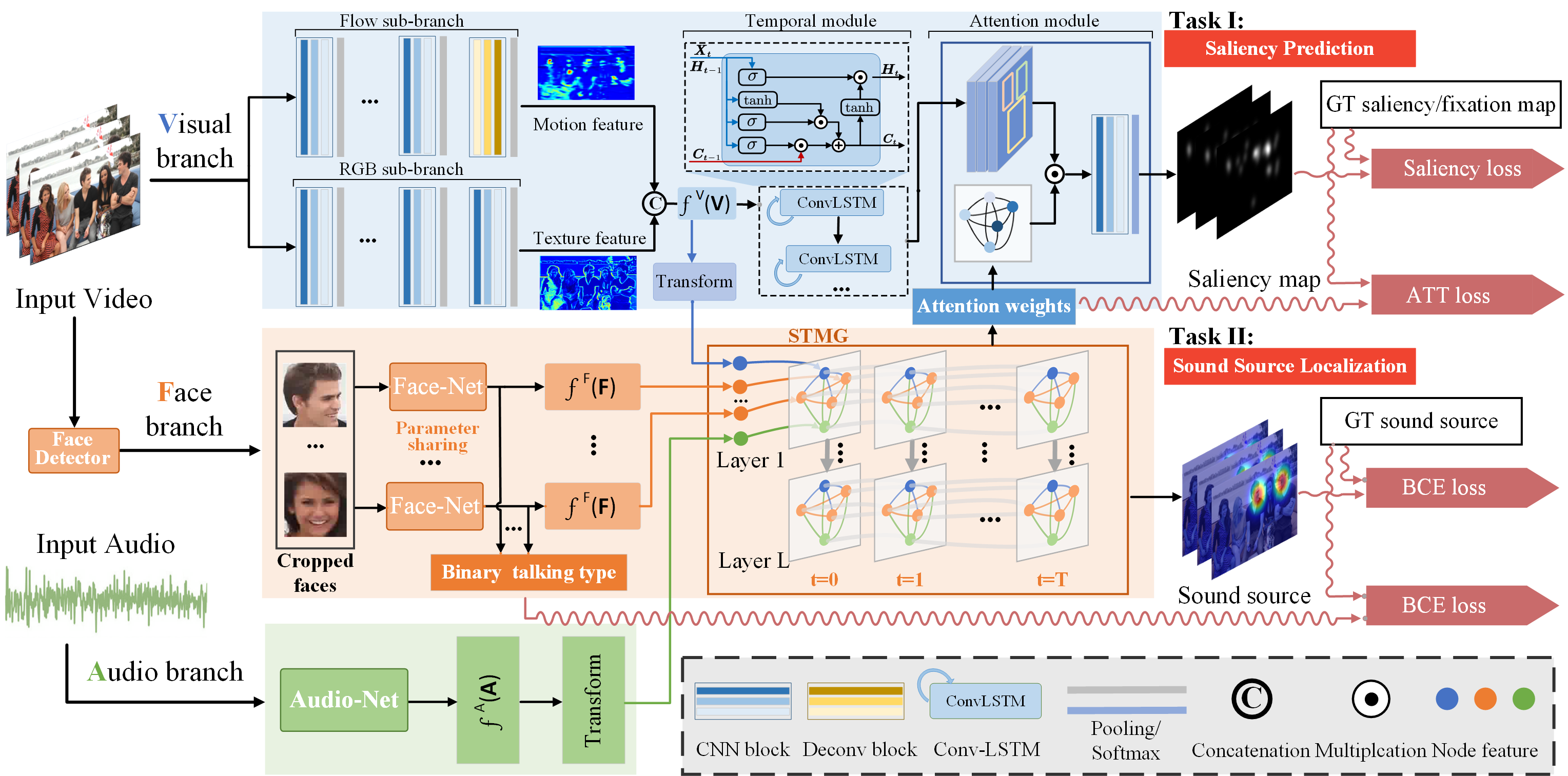}
	\vspace{-1em}
	\hspace{1.6em} 
	\caption{Overall framework of the proposed method. \xma{It includes three branches for the visual, audio and face modalities. 
		}
		\vspace{-1em}
	}
	\label{fig:framework}
\end{figure*}

\xma{
	The overall framework is shown in Fig. \ref{fig:framework}.
	First, a video segment $\mathcal{C}$ = $\{\mathbf{V},  \mathbf{F}, \mathbf{A}\}$, comprising video frames  $\mathbf{V}=$ \\$\{\mathbf{V}_t\}_{t=1}^T$, extracted faces $\mathbf{F}=\{\mathbf{F}_t\}_{t=1}^T$ and audio signal $\mathbf{A}=\{\mathbf{A}_t\}_{t=1}^T$, is fed to VAM-Net\footnote{\xma{Note that the number of face in each video segment is generally consistent across frames, and therefore $\{\mathbf{F}_t\}_{t=1}^T$ are with the same dimension.}}. 
	\xma{Here, $T$ is the total number of frames of the video segment, and $t$ is the frame index.}
}
\xma{
	Subsequently, the visual, face and audio branches encode the input modalities of $\mathbf{V}$, $\mathbf{F}$ and $\mathbf{A}$ into corresponding features, \ie, visual features \mla{$\mathbf{H}_\mathbf{V}=\{\mathbf{h}_{\mathbf{V}_t}\}_{t=1}^T$},
	face features \mla{$\mathbf{H}_\mathbf{F}=\{\mathbf{h}_{\mathbf{F}_t}\}_{t=1}^T$} and audio features. 
	\mla{$\mathbf{H}_\mathbf{A}=\{\mathbf{h}_{\mathbf{A}_t}\}_{t=1}^T$}.
	Then, a spatio-temporal multi-modal graph (STMG) is constructed to fuse these features from three modalitis and to explore the interaction among faces. 
	STMG predicts both the speaking persons to produce sound source maps $\mathbf{M}=\{\mathbf{M}_t\}_{t=1}^T$ and the attention weights to boost saliency prediction. Finally, given \mla{$\mathbf{H}_\mathbf{V}=\{\mathbf{h}_{\mathbf{V}_t}\}_{t=1}^T$} and the attention weights, a temporal and attention module is designed to compute the saliency maps $\mathbf{S}=\{\mathbf{S}_t\}_{t=1}^T$.
	Details about each branch are discussed as follows.
}

\xma{\subsection{Architecture of visual branch} \label{sec:visual_branch}} 

\xma{
	As seen in Fig. \ref{fig:framework}, the visual branch takes video frames  $\mathbf{V}$ (\ie,  $\{\mathbf{V}_t\}_{t=1}^T$) as input, and outputs their saliency maps $\mathbf{S}$ (\ie,  $\{\mathbf{S}_t\}_{t=1}^T$). 
	The visual branch is mainly comprised of the feature extraction module, temporal module and attention module. The feature extraction module 
	aims to encode the visual modality input into visual feature \mla{$\mathbf{H}_{\mathbf{V}}$}. 
	In the visual branch, an RGB sub-branch and a flow sub-branch are constructed to extract texture features and motion features, respectively. 
	\xma{\textit{Finding 4} shows that motion influences attention, and our model thus take into account the motion features. Here, the motion features are directly extracted from the input frames, instead of pre-computed optical flow maps.}
	Note that these two kinds of features have been verified to be effective in predicting video saliency \citep{jiang2021deepvs2}. Then, the extracted features are concatenated, denoted as $\mathrm{C}(\cdot)$, such that the visual features can be obtained as
	\begin{flalign}\label{equ:visual}
		\begin{aligned}
			\mla{\mathbf{H}_{\mathbf{V}}} = \mathrm{C}(g_{\mathrm{RGB}}(\mathbf{V}), g_{\mathrm{OF}}(\mathbf{V})).
		\end{aligned}
	\end{flalign}
	In the above equation, $g_{\mathrm{RGB}}(\cdot)$ represents the RGB sub-branch, consisting of 4 \mla{dilated} CNN blocks of VGG-16 \citep{DBLP:journals/corr/SimonyanZ14a}; $g_{\mathrm{OF}}(\cdot)$ denotes the flow sub-branch, comprising 3 CNN blocks one deconvolutional layer of FlowNet \citep{DFIB15}. 
	Next, the visual feature \mla{$\mathbf{H}_{\mathbf{V}}$} is fed into the temporal module (\ie, a two-layer convolutional LSTM), which is leveraged to process spatio-temporal information. 
	\xma{
		Inspired by \textit{Finding 5} (\ie, the sound source is correlated with fixation distribution), an attention module is devised to incorporate visual features and sound features for saliency prediction.
		In the attention module, it takes advantage of the multi-modal based attention weights $\mathbf{\alpha}^{\text{visual}} = \{\alpha_{nn} \}_{n=1}^{N}$ of $N$ faces generated from STMG. Regrading $\mathbf{\alpha}^{\text{visual}}$ as guidance, 
	}
	the output features from the temporal module are re-weighted and further refined to obtain the final saliency maps $\mathbf{S}$: 
	\begin{flalign}\label{equ:visual_att}
		\begin{aligned}
			\mathbf{S} = \mathrm{Att} (\mathrm{LSTM} (\mla{\mathbf{H}_{\mathbf{V}}}), \bm{\alpha}^{\text{visual}}).
		\end{aligned}
	\end{flalign}
	Here, $\mathrm{Att}(\cdot)$ represents the attention module, which takes advantage of multi-modal information summarised by STMG. It can be formulated as,
	\begin{flalign}\label{equ:visual_att_1}
		\begin{aligned}
			&\mathrm{Att}(\mathbf{I}, \bm{\alpha}^{\text{visual}}) = \\
			&\left\{
			\begin{aligned}
				&g_{\mathrm{conv}} (\mathbf{I}(x,y)\cdot \alpha_{nn}),  (x,y)\in\mathcal{F}^n, n=1,2,...,N \\
				&g_{\mathrm{conv}} (\mathbf{I}(x,y)) \qquad\  ,(x,y)\notin \bigcup\limits_{n=1}^N\mathcal{F}^n,
			\end{aligned}
			\right.
		\end{aligned}
	\end{flalign}
	where $\mathbf{I}$ denotes the input feature of attention module and $\mathbf{I}(x,y)$ is the value of $\mathbf{I}$ at the location of $(x,y)$. Besides, $\mathcal{F}^n$ represents the region of the $n$-th face in $\mathbf{I}$, and $g_{\mathrm{conv}}(\cdot)$ is a 3-layer convolution operation. 
	\xma{
		The details about the parameters of each module are tabulated in Tab. 2 of the supplemental material.
	}
}

\newcolumntype{M}[1]{>{\Centering\arraybackslash}m{#1}}

\mla{\subsection{Architecture of audio branch} \label{sec:audio_branch}} 
\xma{
	\xma{According to \textit{Findings 2, 3} and \textit{6}, audio is essential for both saliency prediction and sound source localization. Thus, we design an audio branch to extract sound related features from audio signals, }
	which is then integrated with other modal features for the tasks of sound source localization and saliency prediction. In particular, the audio branch contains an Audio-Net and a transformation block. Audio-Net adopts SoundNet \citep{aytar2016soundnet} as the backbone, since SoundNet has been demonstrated to be effective in extracting sound related features \citep{pami_avm_8894565}. Specifically, taking the raw audio wave as input, Audio-Net generates a feature vector \mla{$\mathbf{H}_{\mathbf{A}}$} after a sequence of 1-D convolutions and batch normalization \citep{ioffe2015batch}. To agree with the node dimension of STMG, a transformation block consisting of two fully-connected (FC) layers is employed to convert the feature vector 
	into the audio node of STMG, \mla{$\mathbf{\tilde{H}}_{\mathbf{A}} = \{\mathbf{\tilde{h}}_{\mathbf{A}_t}\}_{t=1}^T$}. 
	In summary, our audio branch can be denoted as
	\begin{flalign}\label{equ:visual2}
		\begin{aligned}
			\mla{\mathbf{\tilde{H}}_{\mathbf{A}}} = g_{\rm trans}\{ g_{\rm Audio}(\mathbf{A}) \}.
		\end{aligned}
	\end{flalign}
	In \eqref{equ:visual2}, $g_{\rm audio}(\cdot)$ denotes the function of Aduio-Net, and $g_{\rm trans}(\cdot)$ represents the transformation block in the audio branch.
}

\mla{\subsection{Architecture of the face branch} \label{sec:face_branch}} 

\xma{
	According to \textit{Finding 1}, human attention is more likely to be attracted by one among multiple faces, when  simultaneously presenting audio and video. Therefore, we develop the face branch for localizing the sound source and then providing attention weights to the visual branch for predicting saliency maps. 
	Specifically, the face branch contains a face feature extraction module and a STMG module. It mainly focuses on the task of sound source localization, outputting sound source maps. In particular, as depicted in Fig. \ref{fig:framework}, a face detector \citep{zhang2016joint} is first applied to locate all the faces from the sequence of input video frames. Subsequently, the detected faces are cropped and fed into the face feature module, \ie, Face-Net, to obtain the face features \mla{$\mathbf{H}_{\mathbf{F}}$}. Then, these face features, together with audio feature and visual feature, are represented by the nodes of STMG. The L-layer STMG integrates multi-modal information and exploits the interaction among faces.  
	Finally, STMG outputs the sound classes (\ie, voiced or mute) of each face and the background, \mla{as well as the attention weights for saliency prediction}. %
	Based on the sound classes, the final sound source maps are generated using Gaussian distribution. The details about each part of the face branch are explained as follows.
}

\xma{\noindent{\textbf{Face feature extraction.}} 
	Face feature extraction module, 
	\ie, Face-Net, aims to encode the input of face modality into face feature \mla{$\mathbf{H}_{\mathbf{F}}$} and to preliminarily predict face speaking classes: speaking or non-speaking. Specifically, Face-Net is composed of a convolutional 3D (C3D) model \citep{tran2015learning}, followed by a two-dimensional fully connected (FC) layer. Note that each face corresponds to a Face-Net and the Face-Nets of all input faces share parameters. The 487-dimension features generated from the C3D model are directly fed into STMG, which yields the two-dimensional vectors encapsuling the probabilities of face speaking. 
}

\begin{figure*}[htbp] 
	\centering
	\vspace{-.5em}
	\includegraphics[width=1\linewidth]{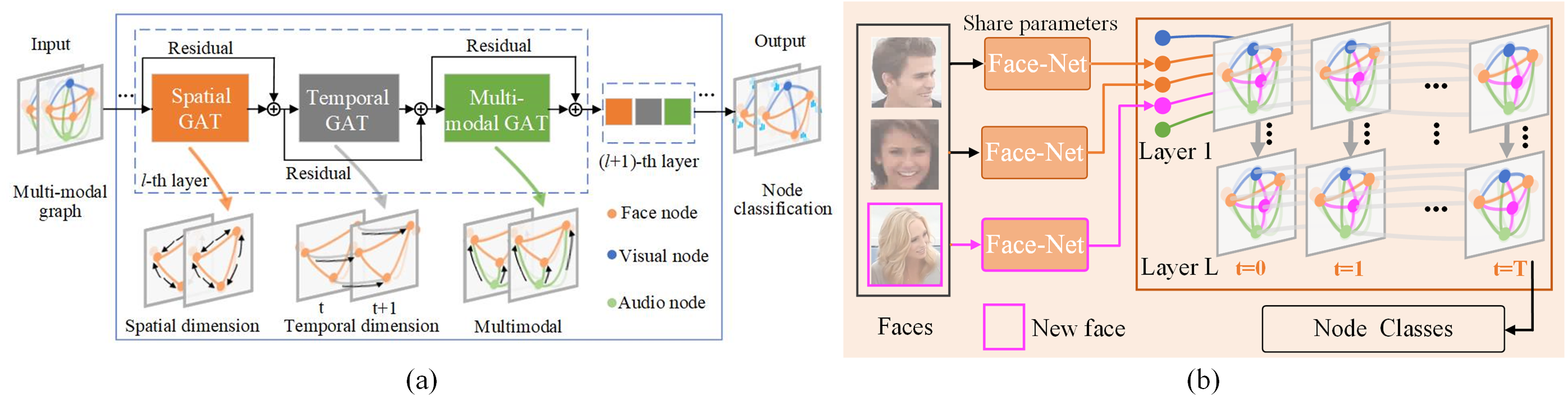}
	\vspace{-2.0em}
	\caption{(a) Structure of the face branch. (b) An example of face branch processing variant face numbers. \mla{Best viewed in colors.
	}}
	\label{fig:face_branch}
	\vspace{-1em}
\end{figure*}

\vspace{-.1em}
\xma{\noindent{\textbf{Graph construction of STMG.}} 
	After obtaining the features of different modalities, we construct STMG $\mathcal{G}(\mathcal{V}, \mathcal{E})$ by utilizing these features as nodes: \mla{$\mathcal{V}=\{ \{\mathbf{h}_{\mathbf{F_t}}^{n}\}_{n=1}^N, \mathbf{\tilde{h}}_{\mathbf{V}_t},$\\$ \mathbf{\tilde{h}}_{\mathbf{A}_t} \}_{t=1}^T$}. Here, \mla{$\mathbf{h}_{\mathbf{F_t}}^{n}$} represents the feature of the $n$-th face at the $t$-th frame. Similarly, \mla{$\mathbf{\tilde{h}}_{\mathbf{V}_t}$} and \mla{$\mathbf{\tilde{h}}_{\mathbf{A}_t}$} are the transformed visual feature and audio feature at the $t$-th frame, respectively.
	Besides, we partition $\mathcal{G}(\mathcal{V}, \mathcal{E})$ into three sub-graphs: spatial graph $\mathcal{G}^S(\mathcal{V}^S, \mathcal{E}^S)$, temporal graph $\mathcal{G}^T(\mathcal{V}^T, \mathcal{E}^T)$ and multi-modal graph $\mathcal{G}^M(\mathcal{V}^M, \mathcal{E}^M)$, 
	As illustrated in Fig \ref{fig:face_branch} (a), in the spatial dimension, the nodes \mla{$\mathcal{V}^S=\{\{\mathbf{h}_{\mathbf{F_t}}^{n}\}_{n=1}^N, \mathbf{\tilde{h}}_{\mathbf{V}_t}\}$} within one frame are fully connected with undirected edges. In the temporal dimension, each spatial node (
	\eg, \mla{$\mathbf{h}_{\mathbf{F_t}}^{n}$}) is forward connected to the same node (\eg, \mla{$\mathbf{h}_{\mathbf{F_{t+1}}}^{n}$}) in the subsequent frame. In the multi-modal dimension, the audio node \mla{$\mathbf{\tilde{h}}_{\mathbf{A}_t}$} is forward connected to each node $\mathcal{V}^S$ in the spatial sub-graph.
}

\xma{\noindent{\textbf{Neural network design of STMG.}}
	\xma{
		With the constructed graph $\mathcal{G}(\mathcal{V}, \mathcal{E})$, a novel STMG neural network is developed to integrate multi-modal information and to learn the spatio-temporal representation.
	}
	Concretely, we start by the computation of a single STMG layer that consists of a spatial graph attention network (GAT), a temporal GAT and a multi-modal GAT in series, as depicted in Fig. \ref{fig:face_branch} (a). The residual is added to each GAT block, in order to construct the deep STMG network. 
	In each GAT, the corresponding sub-graph is updated in a way similar to \citep{velivckovic2017graph}. We take one GAT as an example. Firstly, each node is transformed into an embedding space: 
	\begin{flalign}\label{equ:feature_transfer} 
		\begin{aligned}
			v_{i}^{\prime}=\mathbf{W}v_i,
		\end{aligned}
	\end{flalign}
	where $v_i$ is the feature of the $i$-th node, and $\mathbf{W}$ is a shared linear transformation matrix. Note that $\mathbf{W}$ is shared within the same modal nodes, but it is different cross modalities. 
	Secondly, the transformed features are connected to compute attention coefficients $\mathrm{\alpha}_{ij}$ for each pair of two directly adjacent nodes (\ie, $v_i^{\prime}$ and $v_j^{\prime}$): 
	\begin{flalign}\label{equ:attention_cal} 
		\begin{aligned}
			\mathrm{\alpha}_{ij}=\frac{\exp \left( \mathrm{\sigma}\left( \mathrm{a}^{\mathrm{T}}\left[ \mathrm{W}v_i\parallel \mathrm{W}v_j \right] \right) \right)}{\sum_{k\in \mathcal{K}_i}{\exp \left( \mathrm{\sigma}\left( \mathrm{a}^{\mathrm{T}}\left[ \mathrm{W}v_i\parallel \mathrm{W}v_k \right] \right) \right)}},
		\end{aligned}
	\end{flalign}
	\xma{where $\rm a$ represents the attention vector for computing the importance of one node to another, \eg, $v_i$ to $v_j$, and ${\mathcal{K}_i}$ indicates the neighborhood of node $i$. Note that all node pairs share the same vector $\rm a$.}
	Besides, $\sigma$ is a nonlinear activation function such as leaky rectified linear unit (Leaky ReLU), and $||$ denotes concatenation operation. 
	Afterwards, each node can be updated by fusing the adjacent nodes. In STMG, the multi-head mechanism \citep{NIPS2017_3f5ee243} is adopted to increase its capacity and stability. Thus, the updated node is formulated as:
	\begin{flalign}\label{equ:feature_fusion} 
		\begin{aligned}
			\mathrm{z}_i=\underset{d=1}{\overset{D}{||}}\sigma \left( \sum_{j\in \mathcal{K}_i}{\alpha _{ij}^{d}\mathrm{W}^dv_j} \right).
		\end{aligned} 
	\end{flalign}
	In the above equation, $D$ is the number of heads; $\alpha _{ij}^{d}$ and ${W}^{d}$ are the attention coefficient and transformation matrix of the $d$-th head, respectively. 
	\mla{
		To alleviate the over-smoothing problem of the graph neural
		network, we employ the re-weighted scheme of \citep{ML-GCN_CVPR_2019} to adjust the weight of features for each node and its neighboring nodes.
	}
	For better performance, the updated nodes in the final layer are obtained by computing the average of multi-head results, instead of concatenation in \eqref{equ:feature_fusion}.
}

\xma{
	After a series of STMG layer computations, each face node or the visual node is computed to form a two-dimensional feature vector. 
	By applying the softmax function on the two-dimensional features, we can predict whether each face or background is mute or not. %
	Note that the prediction of the visual node represents the sound class of background.
}

\textbf{Generation of sound source maps.}
Finally, we calculate the sound source map $\mathbf{M}_t$ at the $t$-th frame as follows, 
\begin{flalign}\label{equ:GMM}
	\begin{aligned}
		\mathbf{M}_t = \sum_{n=1}^N \hat{y}_{n,t} \cdot \mathcal{N}_{n,t},
	\end{aligned}
\end{flalign}
where $\hat{y}_{n,t}$ is the predicted sound class of the $n$-th face, \ie, $\hat{y}_{n,t}=1$ represents speaking and $\hat{y}_{n,t}=0$ denotes non-speaking.
Here, we follow \citep{liu2017predicting} to regard the sound source region of the $n$-th face as a Gaussian distribution $\mathcal{N}_{n,t}(\mu_{n,t}, \Sigma_{n,t})$: 
\begin{equation}\label{equ:GMM1}
	\mathcal{N}_{n,t}(\textbf{x})=\mathrm{exp}\{-\frac{1}{2}(\textbf{x}-\bm{\mu}_{n,t})^{\mathrm{T}}\bm{\Sigma}_{n,t}^{-1}(\textbf{x}-\bm{\mu}_{n,t})\},
\end{equation}
\xma{where $\textbf{x}$ indicates the pixel position in 2D space. Additionally, $\bm{\mu}_{n,t}$ means the mean value vector, and $\bm{\Sigma}_{n,t}^{-1}$ represents the covariance matrix. }

\xma{\noindent{\textbf{Processing of variant face number.}} }
\xma{
	As shown in Fig. \ref{fig:face_branch} (b), a new Face-Net and a new node in STMG are instantiated, when a new face appears in the video. The parameter-sharing architecture of Face-Net is able to process videos with variable number of faces. In addition, the GAT-based architecture of STMG allows a new node to be instantiated during computation, since the update of each node does not rely on the number of other nodes.
}

\xma{\noindent{\textbf{Training procedure.}} 
	The overall operation process of STMG network is summarized in Algorithm \ref{alg:multi-modal st-gat}. Given visual, face and audio nodes, STMG is constructed with three sub-graphs: spatial sub-graph $\mathcal{G}^S(\mathcal{V}^S, \mathcal{E}^S)$, temporal sub-graph $\mathcal{G}^T(\mathcal{V}^T, \mathcal{E}^T)$ and multi-modal sub-graph $\mathcal{G}^M(\mathcal{V}^M, \mathcal{E}^M)$. Then, within a layer, the spatial GAT, the temporal GAT and the multi-modal GAT are executed in sequence, using \eqref{equ:feature_transfer}, \eqref{equ:attention_cal} and \eqref{equ:feature_fusion}. Finally, the final layer is computed to obtain the sound class results, as the outputs of the face branch. Besides, the attention coefficient for each node is also output by the face branch, which is fed to the visual branch for saliency prediction.
}

\begin{algorithm}[!t]
	\caption{\mla{Inference scheme of STMG network.}}\label{alg:multi-modal st-gat}%
	\DontPrintSemicolon%
	\SetKw{OR}{or}%
	\SetKw{AND}{and}%
	\KwIn{Features of visual, face and audio nodes at different frames: \mla{$\mathcal{V}=\{ \{\mathbf{h}_{\mathbf{F_t}}^{n}\}_{n=1}^N, \mathbf{\tilde{h}}_{\mathbf{V}_t}, \mathbf{\tilde{h}}_{\mathbf{A}_t} \}_{t=1}^T$}, \hspace{3em} \hspace{3em}
		Graph of spatial, temporal and multi-modal: $
		\mathcal{G}^{\mathrm{S}}(\mathcal{V}^S,\mathcal{E}^S),	\mathcal{G}^{\mathrm{T}}(\mathcal{V}^T,\mathcal{E}^T), \mathcal{G}^{\mathrm{M}}(\mathcal{V}^M,\mathcal{E}^M)$ 
		\hspace{3em} \hspace{3em}
		Number of attention heads: $D$ \hspace{3em} \hspace{3em}
	}
	\KwOut{Talking type of all nodes $y_{n,t}$, \quad \quad \hspace{3em} \hspace{3em} Corresponding predicted attention weights $\bm{\alpha}^\mathrm{Pre}$.}
	Initialize the parameters of $\mathcal{G}^{\mathrm{S}}(\mathcal{V}^S,\mathcal{E}^S)$, $	\mathcal{G}^{\mathrm{T}}(\mathcal{V}^T,\mathcal{E}^T)$, $\mathcal{G}^{\mathrm{M}}(\mathcal{V}^M,$ $\mathcal{E}^M)$, $\bm{\alpha}^\mathrm{Pre} \leftarrow \varnothing$\;
	\For{t = 1 to T}{
		Select the $t$-th spatial graph $\mathcal{G}_t^{\mathrm{S}}(\mathcal{V}_t^S,\mathcal{E}_t^S)$ \\
		\For{d = 1 to D}{
			\For{i = 1 to N+1}{
				Compute attention coefficient $\{{\alpha _{ij}^{d}}\}, j\epsilon \mathcal{K}_i$ for $i$-th face or visual node of $d$-th head at the $t$-th frame, according to  (\ref{equ:feature_transfer}) and (\ref{equ:attention_cal}). \\
			}
		}
		Update each node's feature at the $t$-th frame:
		$\mathrm{z}_i=\underset{d=1}{\overset{D}{||}}\sigma \left( \sum_{j\in \mathcal{K}_i}{\alpha _{ij}^{d}\mathrm{W}^dv_j} \right), i=1, 2, ..., N$.
		Collect attention coefficient for each node:
		
		\vspace{0.2em}
		\hspace{3.2em}
		$\bm{\alpha}^\mathrm{Pre} \gets \,\,\bm{\alpha}^\mathrm{Pre} \bigcup{\{\mathrm{\alpha}_{\mathrm{ii}}^{1}\}_{i=1}^{N+1}}$ \\
	}
	
	\For{t = 2 to T}{
		Select the $t$-th temporal graph $\mathcal{G}_t^{\mathrm{T}}(\mathcal{V}_t^T,\mathcal{E}_t^T)$,\\
		Update each node's feature based on (\ref{equ:feature_transfer}), (\ref{equ:attention_cal}) and (\ref{equ:feature_fusion}).\\ 
	}
	\For{t = 1 to T}{
		Select the $t$-th multi-modal graph
		$\mathcal{G}_t^{\mathrm{M}}(\mathcal{V}_t^M,\mathcal{E}_t^M)$, \\
		Update each faces node's feature using (\ref{equ:feature_transfer}), (\ref{equ:attention_cal}) and (\ref{equ:feature_fusion}). \\
	}
	
	Compute the sound class results $y_{n,t}$ for all face nodes and the visual node. \\
	\Return $y_{n,t}, \bm{\alpha}^\mathrm{Pre}$\;
\end{algorithm}%

\vspace{1.0em}
\subsection{Loss functions and training protocol} \label{sec:lloss_protocol}
\xma{
	Our VAM-Net model aims at solving two tasks: saliency prediction and sound source localization. 
	Accordingly, the optimization of our model can be divided into two parts: the optimization of the sound source localization network and that of the saliency prediction network. 
	The details of each part are discussed in the following.
}

\vspace{0.4em}
\noindent{\xma{\textbf{Loss function of sound source localization.}}}
\xma{
	At each frame, we use both the binary cross entropy (BCE) loss and attention loss (ATT loss) to optimize the performance of sound source localization. In particular, the BCE loss is employed for sound classification (\ie, voiced or mute):
	\begin{flalign}\label{equ:bce_loss} 
		\begin{aligned}
			\mathcal{L}_{\mathrm{BCE}}=&\frac{1}{T}\frac{1}{N+1}\sum_{t=1}^T \sum_{n=1}^{N+1}{\left( \hat{y}_{n,t} \right.}\log \left( {y}_{{n,t}} \right) \\
			& + \left( 1-{\hat{y}}_{n,t} \right) \log \left. \left( 1-{y}_{n,t} \right) \right) ,
		\end{aligned}
	\end{flalign}
	where ${{y}}_{n,t} \in \left\{ 0,1 \right\} $ and $\hat{y}_{n,t} \in \left\{ 0,1 \right\}$ represent GT and predicted binary classes of voiced or mute, respectively, for the $n$-th face at the $t$-th frame. 
}

\xma{In addition, \cite{knyazev2019understanding} found that accurate attention prediction of GNN model can improve the generalization ability and boost the performance in certain classification tasks. Inspired by their finding, we combine an attention loss $\mathcal{L}_{\rm{ATT}}$ using the Kullback-Leibler (KL) divergence for the training process:
}
\begin{flalign}\label{equ:gt_attention} 
	\begin{aligned}
		\xma{
			\mathcal{L}_{\rm{ATT}}=\frac{1}{T}\frac{1}{N+1}\sum_{t=1}^T \sum_{i=1}^{N+1}{\alpha _{ii}^{\rm{GT}}(t)\log \left( \frac{\alpha _{ii}(t)^{\rm{GT}}}{\alpha _{ii}(t)^{\rm{Pre}}} \right)}.
		}
	\end{aligned}
\end{flalign}
\xma{
	In \eqref{equ:gt_attention}, $\alpha _{ii}^{\rm GT}(t)$ and $\alpha _{ii}^{\rm Pre}(t)$ denote the GT and predicted attention values at the $t$-th frame, respectively. Note that $\alpha _{ii}^{\rm GT}$ is regarded as the proportion of fixations falling into certain regions belonging to the $i$-th node.
	Besides, $N$ is the number of face nodes, and $N+1$ indicates the total number of face nodes and the visual node.
	Then, we add the attention loss to the classification loss with ratio $\gamma_1$ to train the STMG network as follows,}
\begin{flalign}\label{equ:nll_loss} 
	\begin{aligned}
		\xma{
			\mathcal{L}_{\mathrm{Sound}} =  \mathcal{L}_{\mathrm{BCE}} + \gamma_1 \cdot \mathcal{L}_{\rm{ATT}}. 
		}
	\end{aligned}
\end{flalign}

\vspace{0.4em}
\noindent{\xma{\textbf{Loss function of saliency prediction.}}}
For saliency prediction on each frame $t$, we use the GT fixation density map $\mathbf{G}_t$ \xma{and fixation location map $\mathbf{P}_t$} to simultaneously supervise the predicted saliency map $\mathbf{S}_t$. 
\xma{Following \citep{wang2018revisiting} and \citep{cornia2018predicting}, we combine three loss functions to train our saliency model:
	\begin{flalign}\label{equ:loss_s}
		\begin{aligned}
			\mathcal{L} _{\text{Saliency}}=\mathcal{L} _{\text{kl}}+\beta _1\mathcal{L} _{\text{nss}}+\beta _2\mathcal{L} _{\text{cc}},
		\end{aligned}
	\end{flalign}
	where $\mathcal{L} _{\text{kl}}$, $\mathcal{L} _{\text{nss}}$ and $\mathcal{L} _{\text{cc}}$ are KL divergence, NSS and CC losses, respectively.
	Moreover, $\beta _1$ and $\beta _2$ are the corresponding weights to balance these three losses.
}

\xma{
	The KL divergence quantifies the distribution difference between the GT and predicted maps, and is computed as follows,
	\begin{flalign}\label{equ:loss_s_kl}
		\begin{aligned}
			\mathcal{L}_{\text{kl}} &= 
			\frac{1}{T} \sum_{t=1}^T \sum_{\textbf{x}\in \mathrm{\mathbf{S}}_t} \mathrm{\mathbf{G}}_t(\textbf{x}) \mathrm{log}\frac{\mathrm{\mathbf{G}}_t(\textbf{x})}{\mathbf{S}_t(\textbf{x})},
		\end{aligned}
	\end{flalign}
	where $\textbf{x}$ denotes the 2D position of each pixel.
}
\xma{
	The NSS loss $\mathcal{L} _{\text{nss}}$ measures the average value of normalized $\mathbf{S}_t$ at GT fixation locations:
	\begin{flalign}\label{equ:loss_nss}
		\begin{aligned}
			\mathcal{L} _{\text{nss}}=\frac{1}{T}\sum_{t=1}^T{\sum_{\textbf{x}\in \mathrm{\mathbf{P}}_{t}}{\frac{\mathbf{S}_{t}\left( \textbf{x} \right) -\mathrm{\mu}\left( \mathbf{S}_{t} \right)}{\mathrm{\sigma}\left( \mathbf{S}_{t} \right)}}}\mathbf{P}_{t}\left( \textbf{x} \right),
		\end{aligned}
	\end{flalign}
	where $\mathrm{\mu}(\cdot)$ and $\mathrm{\sigma}(\cdot)$ indicate the mean and standard deviation, respectively.
}
\xma{
	The CC loss $\mathcal{L} _{\text{cc}}$ evaluates the linear correlation between $\mathbf{S}_t$ and $\mathbf{G}_t$:
	\begin{flalign}\label{equ:loss_cc}
		\begin{aligned}
			\mathcal{L} _{\text{cc}}=\frac{1}{T}\sum_{t=1}^T{\frac{\mathrm{\sigma}\left( \mathbf{S}_{t},\mathbf{G}_{t} \right)}{\mathrm{\sigma}\left( \mathbf{S}_{t} \right) \times \mathrm{\sigma}\left( \mathbf{G}_{t} \right)}},
		\end{aligned}
	\end{flalign}
	where $\mathrm{\sigma}\left( \mathbf{S}_{t},\mathbf{G}_{t} \right)$ is the covariance of $\mathbf{S}_{t}$ and $\mathbf{G}_{t}$.
}

\xma{
	Based on the losses $\mathcal{L}_{\mathrm{Sound}}$ in \eqref{equ:nll_loss} and $\mathcal{L}_{\mathrm{Saliency}}$ in \eqref{equ:loss_s}, the overall loss function for training our VAM-Net is
	\begin{flalign}\label{equ:overall_loss} 
		\begin{aligned}
			\xma{
				\mathcal{L} =  \mathcal{L}_{\mathrm{Saliency}} + \gamma_2 \cdot \mathcal{L}_{\rm{Sound}}, 
			}
		\end{aligned}
	\end{flalign}
	where $\gamma_2$ is a hyper-parameter balancing the saliency prediction loss and the sound localization loss.
}

\xma{
	\xma{Next, we concentrate on the training protocol to optimize the loss function of \eqref{equ:overall_loss}. First, we initialize the visual, face and audio branches with the pre-trained models.}
	In the visual branch, we use the pre-trained parameters of VGG-16 and FlowNet as the initial parameters of the RGB and flow sub-branches, respectively. In the face branch, Face-Net is initialized with the original parameters of C3D, and is then pre-trained utilizing $\mathcal{L}_{\mathrm{BCE}}$ in (\ref{equ:bce_loss}). 
	In the audio branch, Audio-Net is initialized with the parameters of SoundNet \citep{aytar2016soundnet}, which can extract a powerful representation of audio signal.
}
\xma{
	Finally, our VAM-Net is trained via the minimization of the overall loss function \eqref{equ:overall_loss}, such that 
	the tasks of saliency prediction and sound source localization can be jointly learned.
}

\vspace{0.4em}

\section{Experiments and Results}

\begin{table*}
	\centering
	\caption{Accuracy of saliency prediction by our method and 12 competing methods over different databases. \xma{The best scores are marked in \textbf{\xm{bold}}, and the \uline{\ml{underline}} scores indicate the second-best results.}}\label{tab:acc_compare}%
	\vspace{-1.em}
	\resizebox{0.99 \textwidth}{!}{
		\begin{threeparttable}
			\begin{tabular}{c|c|cccccccccccccc}
				\toprule[1.0pt]
				\rowcolor{mygray}
				&\hspace{-0.4em}Metric\hspace{-0.4em}
				&\hspace{-0.4em}Ours\hspace{-0.8em}
				&\hspace{-0.4em}VASM\tnote{1}\hspace{-0.4em}
				&\hspace{-0.4em}TASED\hspace{-0.4em}
				&\hspace{-0.8em}SAM\_res\hspace{-0.5em}
				&\hspace{-0.5em}SAM\_vgg\hspace{-0.5em}
				&\hspace{-0.8em}Liu\hspace{-0.5em}
				&\hspace{-0.8em}ACLNet\hspace{-0.5em}
				&\hspace{-0.4em}DeepVS \hspace{-0.4em}
				&\hspace{-0.8em}SalGAN\hspace{-0.5em}
				&\hspace{-0.8em}Coutrot\hspace{-0.4em}
				&\hspace{-0.8em}SALICON\hspace{-0.5em}
				&\hspace{-0.8em}OBDL\hspace{-0.5em}
				&\hspace{-0.8em}BMS\hspace{-0.8em}
				&\hspace{-0.8em}G-Eymol\hspace{-0.5em}\\
				
				\midrule
				\midrule
				
				\multirow{4}{*}{\tabincell{c}{\hspace{-0.8em}\rotatebox{90}{MVVA}\hspace{-0.8em}}}\hspace{-0.8em}
				&\hspace{-0.8em}AUC\hspace{-0.8em}
				&\hspace{-0.4em}\textbf{\xm{0.912}}\hspace{-0.8em}
				&\hspace{-0.4em}\uline{\ml{0.905}}\hspace{-0.8em}
				&\hspace{-0.8em}\uline{\ml{0.905}}\hspace{-0.8em}
				&\hspace{-0.8em}{0.897}\hspace{-0.8em}
				&\hspace{-0.8em}0.896\hspace{-0.8em}
				&\hspace{-0.8em}0.893\hspace{-0.8em}
				&\hspace{-0.8em}0.889\hspace{-0.8em}
				&\hspace{-0.8em}0.890\hspace{-0.8em}
				&\hspace{-0.8em}0.891\hspace{-0.8em}
				&\hspace{-0.8em}0.869\hspace{-0.8em}
				&\hspace{-0.8em}0.866\hspace{-0.8em}
				&\hspace{-0.8em}0.786\hspace{-0.8em}
				&\hspace{-0.8em}0.765\hspace{-0.8em}
				&\hspace{-0.8em}0.615\hspace{-0.5em}\\
				
				&\hspace{-0.8em}NSS\hspace{-0.8em}
				&\hspace{-0.4em}\textbf{\xm{4.002}}\hspace{-0.8em}
				&\hspace{-0.4em}\uline{\ml{3.976}}\hspace{-0.8em}
				&\hspace{-0.8em}3.319\hspace{-0.8em}
				&\hspace{-0.8em}{3.495}\hspace{-0.8em}
				&\hspace{-0.8em}3.466\hspace{-0.8em}
				&\hspace{-0.8em}3.279\hspace{-0.8em}
				&\hspace{-0.8em}3.437\hspace{-0.8em}
				&\hspace{-0.8em}3.270\hspace{-0.8em}
				&\hspace{-0.8em}2.650\hspace{-0.8em}
				&\hspace{-0.8em}2.604\hspace{-0.8em}
				&\hspace{-0.8em}2.523\hspace{-0.8em}
				&\hspace{-0.8em}1.342\hspace{-0.8em}
				&\hspace{-0.8em}0.936\hspace{-0.8em}
				&\hspace{-0.8em}0.551\hspace{-0.5em}\\
				
				&\hspace{-0.8em}CC\hspace{-0.8em}
				&\hspace{-0.4em}\textbf{\xm{0.741}}\hspace{-0.8em}
				&\hspace{-0.4em}\uline{\ml{0.722}}\hspace{-0.8em}
				&\hspace{-0.8em}{0.653}\hspace{-0.8em}
				&\hspace{-0.8em}0.634\hspace{-0.8em}
				&\hspace{-0.8em}0.634\hspace{-0.8em}
				&\hspace{-0.8em}0.625\hspace{-0.8em}
				&\hspace{-0.8em}0.639\hspace{-0.8em}
				&\hspace{-0.8em}0.615\hspace{-0.8em}
				&\hspace{-0.8em}0.539\hspace{-0.8em}
				&\hspace{-0.8em}0.509\hspace{-0.8em}
				&\hspace{-0.8em}0.477\hspace{-0.8em}
				&\hspace{-0.8em}0.273\hspace{-0.8em}
				&\hspace{-0.8em}0.193\hspace{-0.8em}
				&\hspace{-0.8em}0.125\hspace{-0.5em}\\
				
				&\hspace{-0.8em}KL\hspace{-0.8em}
				&\hspace{-0.4em}\textbf{\xm{0.783}}\hspace{-0.8em}
				&\hspace{-0.4em}\uline{\ml{0.823}}\hspace{-0.8em}
				&\hspace{-0.8em}{0.970}\hspace{-0.8em}
				&\hspace{-0.8em}1.004\hspace{-0.8em}
				&\hspace{-0.8em}1.012\hspace{-0.8em}
				&\hspace{-0.8em}1.098\hspace{-0.8em}
				&\hspace{-0.8em}1.044\hspace{-0.8em}
				&\hspace{-0.8em}1.117\hspace{-0.8em}
				&\hspace{-0.8em}1.234\hspace{-0.8em}
				&\hspace{-0.8em}1.557\hspace{-0.8em}
				&\hspace{-0.8em}1.447\hspace{-0.8em}
				&\hspace{-0.8em}1.995\hspace{-0.8em}
				&\hspace{-0.8em}2.051\hspace{-0.8em}
				&\hspace{-0.8em}4.253\hspace{-0.5em}\\
				
				\midrule
				\midrule
				
				\multirow{4}{*}{\tabincell{c}{\rotatebox{90}{\vspace{-1.8em}  Coutrot II}}}
				&\hspace{-0.8em}AUC\hspace{-0.5em}
				&\hspace{-0.4em}\textbf{\xm{0.925}}\hspace{-0.8em}
				&\hspace{-0.4em}\uline{\ml{0.922}}\hspace{-0.5em}
				&\hspace{-0.8em}0.877\hspace{-0.8em}
				&\hspace{-0.8em}0.905\hspace{-0.5em}
				&\hspace{-0.8em}0.849\hspace{-0.5em}
				&\hspace{-0.8em}{0.908}\hspace{-0.5em}
				&\hspace{-0.8em}0.848\hspace{-0.8em}
				&\hspace{-0.8em}0.896\hspace{-0.8em}
				&\hspace{-0.8em}0.900\hspace{-0.8em}
				&\hspace{-0.8em}0.883\hspace{-0.8em}
				&\hspace{-0.8em}0.865\hspace{-0.8em}
				&\hspace{-0.8em}0.723\hspace{-0.8em}
				&\hspace{-0.8em}0.751\hspace{-0.8em}
				&\hspace{-0.8em}0.698\hspace{-0.5em}\\
				
				&\hspace{-0.8em}NSS\hspace{-0.5em}
				&\hspace{-0.4em}\textbf{\xm{3.682}}\hspace{-0.8em}
				&\hspace{-0.4em}\uline{\ml{3.568}}\hspace{-0.5em}
				&\hspace{-0.8em}2.731\hspace{-0.8em}
				&\hspace{-0.8em}{3.446}\hspace{-0.5em}
				&\hspace{-0.8em}3.306\hspace{-0.5em}
				&\hspace{-0.8em}2.833\hspace{-0.5em}
				&\hspace{-0.8em}3.127\hspace{-0.5em}
				&\hspace{-0.8em}3.058\hspace{-0.5em}
				&\hspace{-0.8em}2.286\hspace{-0.5em}
				&\hspace{-0.8em}3.033\hspace{-0.8em}
				&\hspace{-0.8em}2.408\hspace{-0.5em}
				&\hspace{-0.8em}0.730\hspace{-0.8em}
				&\hspace{-0.8em}0.739\hspace{-0.8em}
				&\hspace{-0.8em}0.884\hspace{-0.5em}\\
				
				&\hspace{-0.8em}CC\hspace{-0.5em}
				&\hspace{-0.4em}\textbf{\xm{0.665}}\hspace{-0.8em}
				&\hspace{-0.4em}\uline{\ml{0.639}}\hspace{-0.5em}
				&\hspace{-0.8em}0.545\hspace{-0.8em}
				&\hspace{-0.8em}{0.607}\hspace{-0.5em}
				&\hspace{-0.8em}0.593\hspace{-0.5em}
				&\hspace{-0.8em}0.585\hspace{-0.5em}
				&\hspace{-0.8em}0.521\hspace{-0.5em}
				&\hspace{-0.8em}0.556\hspace{-0.5em}
				&\hspace{-0.8em}0.553\hspace{-0.5em}
				&\hspace{-0.8em}0.606\hspace{-0.8em}
				&\hspace{-0.8em}0.433\hspace{-0.5em}
				&\hspace{-0.8em}0.181\hspace{-0.8em}
				&\hspace{-0.8em}0.153\hspace{-0.8em}
				&\hspace{-0.8em}0.162\hspace{-0.5em}\\
				
				&\hspace{-0.8em}KL\hspace{-0.5em}
				&\hspace{-0.4em}\uline{\ml{0.984}}\hspace{-0.8em}
				&\hspace{-0.4em}\textbf{\xm{0.915}}\hspace{-0.5em}
				&\hspace{-0.8em}1.271\hspace{-0.8em}
				&\hspace{-0.8em}{1.031}\hspace{-0.5em}
				&\hspace{-0.8em}1.093\hspace{-0.5em}
				&\hspace{-0.8em}1.035\hspace{-0.5em}
				&\hspace{-0.8em}1.357\hspace{-0.5em}
				&\hspace{-0.8em}1.209\hspace{-0.5em}
				&\hspace{-0.8em}1.717\hspace{-0.5em}
				&\hspace{-0.8em}1.428\hspace{-0.8em}
				&\hspace{-0.8em}1.514\hspace{-0.5em}
				&\hspace{-0.8em}2.228\hspace{-0.4em}
				&\hspace{-0.2em}2.073\hspace{-0.8em}&\hspace{-0.8em}2.932\hspace{-0.5em}\\
				
				\midrule
				\midrule
				
				\multirow{4}{*}{\tabincell{c}{\rotatebox{90}{{\ \ \ \ Coutrot III}}}}
				&\hspace{-0.8em}AUC\hspace{-0.5em}
				&\hspace{-0.4em}\uline{\ml{0.927}}\hspace{-0.8em}
				&\hspace{-0.4em}0.925\hspace{-0.5em}
				&\hspace{-0.8em}0.910\hspace{-0.8em}
				&\hspace{-0.8em}\textbf{\xm{0.933}}\hspace{-0.5em}
				&\hspace{-0.8em}\textbf{\xm{0.933}}\hspace{-0.5em}
				&\hspace{-0.8em}0.902\hspace{-0.5em}
				&\hspace{-0.8em}0.918\hspace{-0.5em}
				&\hspace{-0.8em}0.914\hspace{-0.5em}
				&\hspace{-0.8em}0.92\hspace{-0.5em}
				&\hspace{-0.8em}0.904\hspace{-0.5em}
				&\hspace{-0.8em}0.889\hspace{-0.8em}
				&\hspace{-0.8em}0.826\hspace{-0.8em}
				&\hspace{-0.8em}0.632\hspace{-0.5em}
				&\hspace{-0.8em}0.740\hspace{-0.5em}\\
				
				&\hspace{-0.8em}NSS\hspace{-0.5em}
				&\hspace{-0.4em}\textbf{\xm{4.609}}\hspace{-0.8em}
				&\hspace{-0.4em}\uline{\ml{4.032}}\hspace{-0.5em}
				&\hspace{-0.8em}{3.224}\hspace{-0.8em}
				&\hspace{-0.8em}{3.569}\hspace{-0.5em}
				&\hspace{-0.8em}{3.310}\hspace{-0.5em}
				&\hspace{-0.8em}{2.565}\hspace{-0.5em}
				&\hspace{-0.8em}{2.873}\hspace{-0.5em}
				&\hspace{-0.8em}{3.804}\hspace{-0.5em}
				&\hspace{-0.8em}{3.009}\hspace{-0.5em}
				&\hspace{-0.8em}3.028\hspace{-0.5em}
				&\hspace{-0.8em}{2.458}\hspace{-0.8em}
				&\hspace{-0.8em}{1.646}\hspace{-0.8em}
				&\hspace{-0.8em}{0.216}\hspace{-0.5em} 
				&\hspace{-0.8em}{1.010}\hspace{-0.5em} \\
				
				&\hspace{-0.8em}{CC}\hspace{-0.5em}
				&\hspace{-0.4em}\textbf{\xm{0.566}}\hspace{-0.8em}
				&\hspace{-0.4em}\uline{\ml{0.474}}\hspace{-0.5em}
				&\hspace{-0.8em}{0.442}\hspace{-0.8em}
				&\hspace{-0.8em}{0.459}\hspace{-0.5em}
				&\hspace{-0.8em}{0.442}\hspace{-0.5em}
				&\hspace{-0.8em}{0.365}\hspace{-0.5em}
				&\hspace{-0.8em}{0.413}\hspace{-0.5em}
				&\hspace{-0.8em}{0.467}\hspace{-0.5em}
				&\hspace{-0.8em}{0.434}\hspace{-0.5em}
				&\hspace{-0.8em}0.349\hspace{-0.5em}
				&\hspace{-0.8em}{0.292}\hspace{-0.8em}
				&\hspace{-0.8em}{0.252}\hspace{-0.8em}
				&\hspace{-0.8em}{0.031}\hspace{-0.5em} 
				&\hspace{-0.8em}{0.254}\hspace{-0.5em}\\
				
				&\hspace{-0.8em}{KL}\hspace{-0.5em}
				&\hspace{-0.4em}\uline{\ml{1.382}}\hspace{-0.8em}
				&\hspace{-0.4em}\textbf{\xm{1.375}}\hspace{-0.5em}
				&\hspace{-0.8em}{1.584}\hspace{-0.8em}
				&\hspace{-0.8em}{1.440}\hspace{-0.5em}
				&\hspace{-0.8em}{1.479}\hspace{-0.5em}
				&\hspace{-0.8em}{1.905}\hspace{-0.5em}
				&\hspace{-0.8em}{1.546}\hspace{-0.5em}
				&\hspace{-0.8em}{1.689}\hspace{-0.5em}
				&\hspace{-0.8em}{1.606}\hspace{-0.5em}
				&\hspace{-0.8em}{2.111}\hspace{-0.5em}
				&\hspace{-0.8em}{2.145}\hspace{-0.8em}
				&\hspace{-0.8em}{2.276}\hspace{-0.8em}
				&\hspace{-0.8em}{2.770}\hspace{-0.5em} 
				&\hspace{-0.8em}{2.376}\hspace{-0.5em}\\
				\bottomrule[1.0pt]
			\end{tabular}%
			
			\begin{tablenotes}
				\footnotesize
				\item[1] \xma{VASM is the  method of our conference paper.}
			\end{tablenotes}
	\end{threeparttable}}
\end{table*}%

\subsection{Settings}
\yfa{\noindent{\textbf{Configuration.}}} 
\xma{
	In our experiments, our MVVA database is randomly divided into training (240 videos) and test (60 videos) sets. For training and inference, the configuration of our VAM-Net model is described as follows.
	For the visual branch, the input RGB frames are resized to $256\times256$. Then, the resized frames are fed into the RGB sub-branch, and every pair of two consecutive frames with interval of 5 frames is fed into the Flow sub-branch.
	To train the convolutional LSTM, we temporally segment 240 training videos into 5,747 clips, all of which contain $T = 20$ frames. 
}
\xma{For the audio branch, the raw audio wave is re-sampled at rate of 22,050 Hz. Subsequently, we crop 10 seconds of the audio data centered in the middle time of each batch of visual frames.}
For the face branch, the resolution of $N$ input faces is $112 \times 112$.
The parameters of the proposed VAM-Net are updated by using the Stochastic Gradient Descent (SGD) algorithm with Adam optimizer. 
\xma{
	In addition, the key hyper-parameters for training VAM-Net are listed in \mla{Tab. 3 of supplemental material}.
	\mla{All experiments are conducted on a computer with Intel(R) Xeon(R) E5-2698 CPU @2.20GHz, 252 GB RAM and 4 Nvidia Tesla V100 GPUs.}
}

\xma{\noindent{\textbf{Evaluation metrics.}}} 
\xma{
	To evaluate the performance of saliency prediction, 
	we adopt four widely used metrics: area under the receiver operating characteristic curve (AUC), NSS, linear correlation coefficient (CC), and KL divergence. 
	The former two metrics are location based metrics, while the last two are distribution based ones. Note that the larger values for AUC, NSS or CC indicate more accurate saliency prediction, and the opposite holds for the KL divergence. Refer to \citep{bylinskii2018different} for more details on these metrics.
}
\xma{For evaluating sound source localization, four metrics are employed, \ie, the accuracy of predicted sound class (Acc), intersection over union (IoU), \xma{AUC for sound source localization (AUC-S)} and the mean average precision (mAP) \citep{roth2020ava}. For the Acc value, we compute the percentage of correctly predicted sound classification for each test video. Besides, the computation of IoU and \xma{AUC-S} follows \citep{pami_avm_8894565}. Specifically, we first generate the GT binary sound source map $\mathbf{Y}_t$ according to the talking-face box. The value of $\mathbf{Y}_t$ inside the talking-face bounding box is set to be 1; otherwise, it is set to be 0. Recall that $\mathbf{M}_t$ is sound source map at frame $t$. Here, $\mathbf{M}_t$ is binarized by a threshold value, denoted as $\mathbf{\bar{M}}_t$. Then, the IoU can be calculated by
	\begin{flalign}\label{equ:iou} 
		\begin{aligned}
			\mathrm{IoU}=\frac{\mathrm{R}\left( \mathbf{Y}_{\mathrm{t}}\bigcap{\mathbf{\bar{M}}_{\mathrm{t}}} \right)}{\mathrm{R}\left( \mathbf{Y}_{\mathrm{t}}\bigcup{\mathbf{\bar{M}}_{\mathrm{t}}} \right)},
		\end{aligned}
	\end{flalign}
	where $\mathrm{R}\left( \cdot \right)$ is the summed value of the binary map. Finally, the \xma{AUC-S} is obtained as the area under receiver operating characteristic (ROC) curve, 
	with varying the IoU at different thresholds.
}

\subsection{Performance Comparison}
\vspace{-1em}
\xma{\subsubsection{Evaluation of saliency prediction}}
\xma{Here,} we compare the performance of our multi-modal method with \xma{12 state-of-the-art saliency prediction methods}, including TASED \citep{min2019tased}, SAM \citep{cornia2018predicting}, \xma{VASM \mla{(our conference paper)} \citep{ml_salient_face}, Liu \citep{liu2017predicting}}, ACLNet \citep{wang2018revisiting}, DeepVS \citep{jiang2021deepvs2}, SalGAN \citep{pan2017salgan}, SALICON \citep{huang2015salicon}, Coutrot \citep{coutrot2015efficient}, OBDL \citep{hossein2015many}, BMS \citep{zhang2016exploiting} and G-Eymol \citep{zanca2019gravitational}. Among them, SalGAN, SALICON, SAM and BMS are state-of-the-art saliency prediction methods for images, and others are for videos.
\xma{Coutrot, Liu and VASM focus on saliency prediction on multi-face videos.}
In our experiments, we compare two versions of SAM, SAM\_res with the ResNet backbone and SAM\_vgg with the VGGNet backbone. 

\begin{figure*}[htb]
	\begin{center}
		\vspace{-1em}
		\includegraphics[width=1\linewidth]{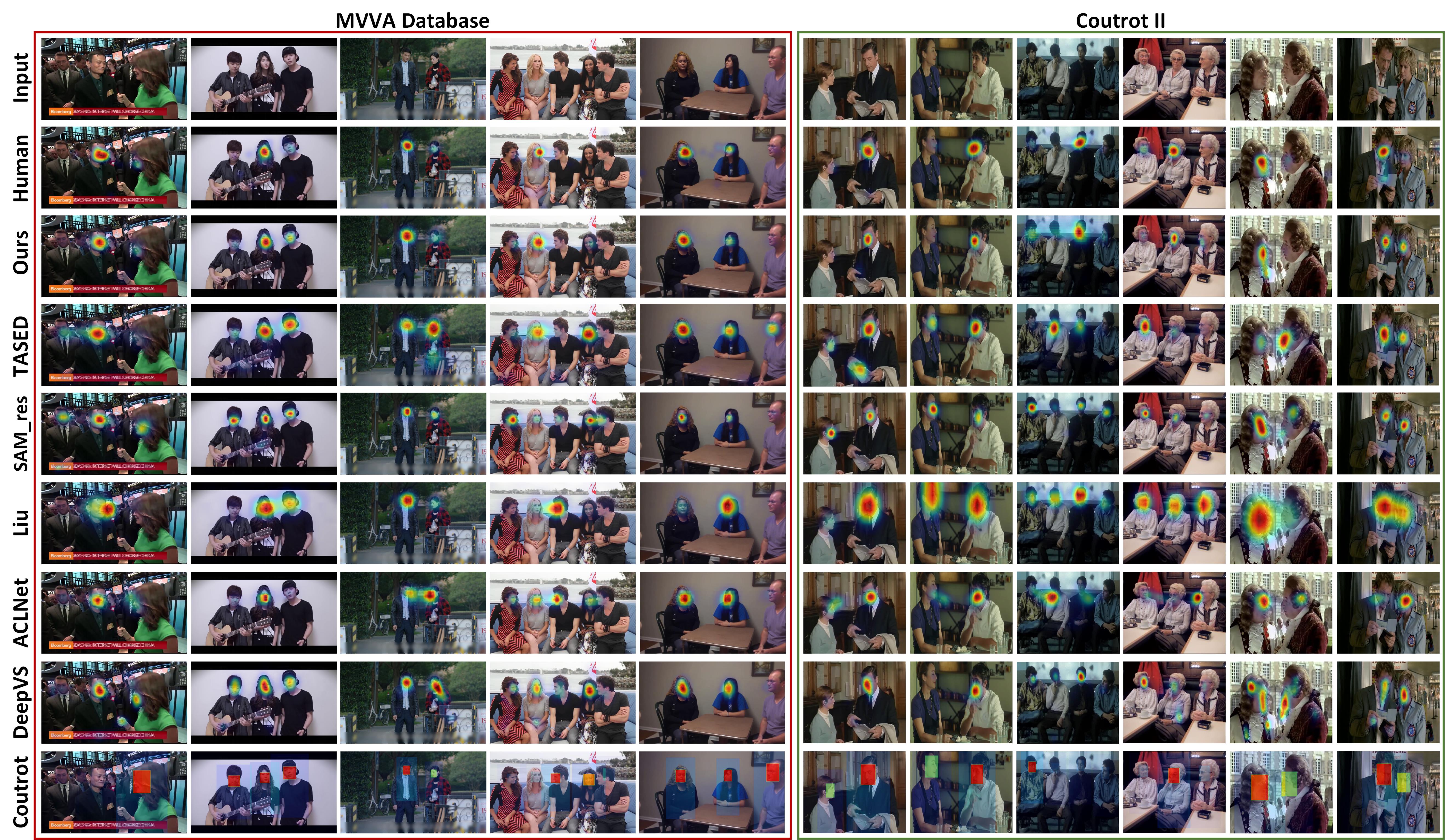}
	\end{center}
	\vspace{-1.0em}
	\caption{\xma{Saliency maps of 11 videos randomly selected from the test set of our eye-tracking database and Coutrot II \citep{coutrot2014saliency}. These qualitative results are generated by our method and other 6 compared approaches, including TASED \citep{min2019tased}, SAM \citep{cornia2018predicting}, Liu \citep{liu2017predicting}, DeepVS \citep{jiang2021deepvs2}, ACLNet \citep{wang2018revisiting} and G-Eymol \citep{zanca2019gravitational}. More results are presented in the supplemental material.}}
	\label{fig:vis_result1}
	\vspace{-1em}
\end{figure*}

\begin{figure*}[t]
	\begin{center}
		\includegraphics[width=1\linewidth]{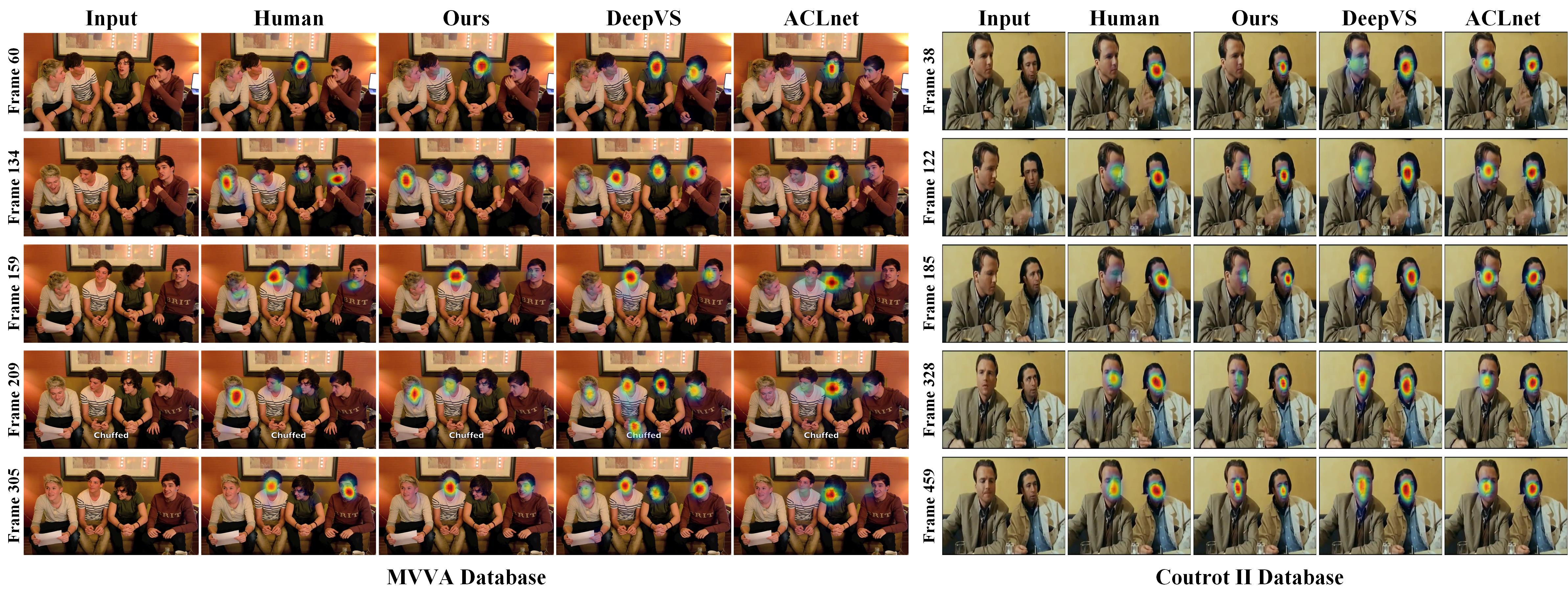}
		\vspace{-1em}
		\caption{Saliency maps for different frames of two video sequences, selected from our MVVA and Coutrot II \citep{coutrot2014saliency}.}
		\label{fig:vis_result2}
	\end{center}
	\vspace{-2.0em}
\end{figure*}

\noindent\textbf{Evaluation on our database.}
Tab. \ref{tab:acc_compare} presents the results of AUC, NSS, CC and KL divergence, \xma{which are averaged over 60 test videos in our eye-tracking database, for our and other methods}. As shown in this table, the proposed method performs significantly better than all other methods in terms of all 4 metrics.
In particular, \xma{our method improves NSS, CC and KL by 0.026, 0.019 and 0.04 over the second best method}. \xma{The main reasons for the improvement are: 1) Most of the state-of-the-art methods do not consider audio information, while our method utilizes the audio cue for saliency prediction, 2) The face subnet of our method learns the face-related features to predict salient faces, and 3) Our STMG effectively integrates the multi-modal information and sufficiently explores the interaction among multiple faces for saliency prediction.} 
Fig. \ref{fig:vis_result1} shows the saliency maps of some randomly selected videos, which are predicted by the proposed method and 12 other methods.
As seen in this figure, our method is capable of precisely locating the salient faces, much closer to the GT.
Fig. \ref{fig:vis_result2} further shows the saliency maps of the successive frames of a selected video.
We can see that our method is also able to precisely predict attention transition across faces, considerably better than other methods.

\noindent\textbf{Evaluation on generalization ability.}
To evaluate the generalization ability of the proposed method, we further evaluate our method and \xma{12 other methods on the Coutrot II \citep{coutrot2014saliency} and Coutrot III \citep{coutrot2015efficient} databases.} 
As shown in Tab. \ref{tab:acc_compare}, the proposed method again outperforms all other methods. 
In particular, we gain at least 0.026 (\mla{0.092}) and 0.114 (\mla{0.577}) improvements in CC and NSS \mla{on Coutrot II (Coutrot III)}, respectively.
Fig. \ref{fig:vis_result1} and \ref{fig:vis_result2} show the saliency maps of some selected videos.
We can see from these figures that our method outperforms other methods in predicting saliency maps and saliency transition across frames.

\begin{table}[!t]
	\centering
	\caption{\xma{Performance of different sound source localization methods on our MVVA database.}}
	\vspace{-1em}
	\resizebox{1\linewidth}{!}{%
	\begin{tabular}{c|lcccc} %
			\toprule[1.0pt]
			\rowcolor{mygray}
			\textbf{Ouput} & \textbf{Method} & \textbf{IoU} & \textbf{AUC-S} & \textbf{Acc} & \textbf{mAP} \\
			\midrule
			\multirow{3}{*}{\tabincell{c}{Sound \\ source \\ map}} & Owens \etal (MSE) & \mla{37.85} & \mla{\textbf{53.22}}  & \mla{71.74} & \mla{58.26} \\ 
			& Tian \etal (AVE) &  \mla{37.80}   & \mla{28.11} & \mla{59.47} & \mla{47.69} \\
			& Senocak \etal (AVM) & \mla{40.10}  & \mla{29.77} & \mla{62.51} & \mla{49.05} \\
			\midrule
			\midrule
			\rowcolor{mygray}
			\textbf{Ouput} & \textbf{Method} & \textbf{IoU} & \textbf{AUC-S} & \textbf{Acc} & \textbf{mAP}\\
			\midrule
			\multirow{2}{*}{\tabincell{c}{Speaking \\ class}}  & Alcázar \etal (ASC) & 24.93  & 23.53 & \mla{67.14} & \mla{68.98} \\   
			& \textbf{Ours (STMG Network)} & \textbf{52.01} & {42.84} & \textbf{78.49} & \textbf{74.22}  \\
			\bottomrule[1.0pt]
	\end{tabular}
}
\label{tab:sound_result}
\vspace{-1em}
\end{table}

\xma{\subsubsection{Evaluation of sound source localization.}}
\xma{
	For sound source localization, we compare our VAM-Net with 4 sound source localization approaches, including AVE\citep{tian2018audio}, MSE\citep{owens2018audio}, VAM \citep{pami_avm_8894565} and ASC \citep{alcazar2020active}. 
	Among them, VAM \citep{pami_avm_8894565} is an image based method, while AVE\citep{tian2018audio} and MSE\citep{owens2018audio} are designed for videos. These three methods all generate confidence maps of sound sources, but ASC \citep{alcazar2020active} predicts the speaking classes (\ie, speaking or non-speaking) of different speakers.
	In contrast, our method can output both speaking classes and sound source maps through (\ref{equ:GMM}) and (\ref{equ:GMM1}).
	For fairness of the performance comparison, we uniformly use IoU, AUC-S, Acc and mAP to evaluate our and other methods. Specifically, to compute Acc and mAP of MSE, AVE and AVM, we convert the predicted confidence maps into bounding boxes, by introducing the prior of face positions. The face bounding box region that has the highest confidence value is regarded as the speaking person.
	To compute IoU and AUC-S of ASC, we convert the predicted speaking bounding boxes into confidence maps with the binary values of $\{0, 1\}$, indicating whether a face is speaking or not.
}

\begin{table}[!t]
	\centering
	\caption{Performance of different modules in our model.}
	\vspace{-1em}
	\resizebox{0.48 \textwidth}{!}{
		\begin{tabular}{c|cccc}
			\toprule[1.0pt]
			\rowcolor{mygray}
			Models & CC & KL & NSS & AUC \\
			\hline
			\hline
			Avg. baseline & 0.364 & 1.575 & 1.614 & 0.848 \\
			\hline
			visual (RGB+flow) & 0.682 & 0.933 & 3.713 & 0.901 \\
			visual (RGB+flow+LSTM) & 0.702 & 0.925 & 3.743 & 0.911 \\
			\yfa{visual+face}  & \mla{0.725} & \mla{0.834} & \mla{3.972} & \mla{0.903} \\
			\yfa{visual+face+audio} & \mla{\textbf{0.741}} & \mla{\textbf{0.783}} & \mla{\textbf{4.002}} & {\mla{\textbf{0.912}}}
			\\
			\hline
			Human & 0.747 & 1.278 & 4.573 & 0.875 \\
			\bottomrule[1.0pt]
	\end{tabular}}%
	\label{tab:ablation}%
		\vspace{-1.5em}
\end{table}%

\xma{
	The quantitative results of our VAM-Net method and other 4 state-of-the-art methods over the MVVA database are reported in Tab. \ref{tab:sound_result}. 
	It can be seen that our VAM-Net method achieves significant improvement over all compared methods in terms of IoU, Acc and mAP metrics.
	In particular, the proposed VAM-Net gains 6.75 improvement in Acc and 15.96 improvement in mAP, over the second best method (MSE). 
	The main reason for such improvements lies in that the proposed method can better mine the correlation among the audio, visual and face modalities, and is promoted by the saliency prediction task.
}

\begin{figure}[t]
	\begin{center}
		\includegraphics[width=1.0\linewidth]{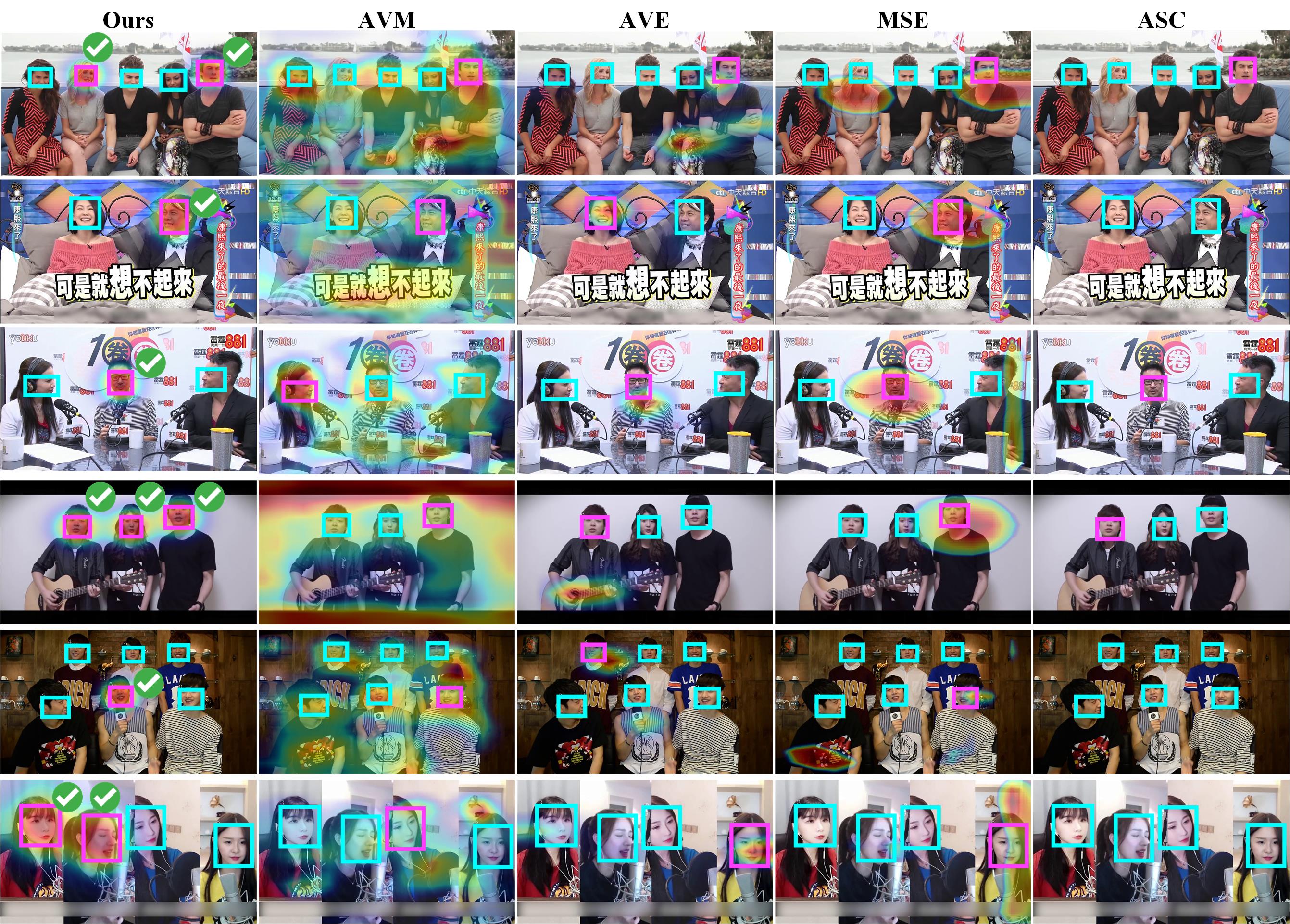}
	\end{center}
	\vspace{-1.0em}
	\caption{\xma{Qualitative results of sound source localization predicted by different approaches on our MVVA database. The heat maps show the confidence of sound source, while the magenta and blue bounding boxes illustrate talking faces and non-talking faces, respectively. The green checkmark means the GT of talking faces.}} 
	\label{fig:sound1_multi_frame}
	\vspace{-1.5em}
\end{figure}

\xma{
	We further compare the qualitative results of our method and compared methods in Fig. \ref{fig:sound1_multi_frame}.
	As can be seen in this figure, our method accurately locates sound source regions, while other methods often wrongly predict the sound regions. 
	Hence, these qualitative results again indicate that our method is more effective in sound source localization of multi-face videos, significantly better than other state-of-the-art methods.
}

\vspace{15pt}
\subsection{Ablation Analysis}
Here, we thoroughly analyze the effectiveness of each module in the proposed method.

\begin{table}[t]
	\centering
	\caption{\xma{Performance of face branch with different components.}}
	\vspace{-1em}
	\resizebox{0.49 \textwidth}{!}{
		\begin{tabular}{ccccc|cc}
			\toprule[1.0pt]
			\rowcolor{mygray}
			\multicolumn{5}{c|}{Components}       & \multicolumn{2}{c}{Metric} \\
			\midrule
			\rowcolor{mygray}
			Face-Net  & spatial & temporal & audio & visual & Acc(\%)\\
			\midrule
			\CheckmarkBold & \XSolidBrush & \XSolidBrush & \XSolidBrush & \XSolidBrush &   \mla{77.06} \\
			\CheckmarkBold & \CheckmarkBold & \XSolidBrush & \XSolidBrush & \XSolidBrush &   \mla{77.33} \\
			\CheckmarkBold & \CheckmarkBold & \CheckmarkBold & \XSolidBrush & \XSolidBrush & \mla{77.45} \\
			\CheckmarkBold & \CheckmarkBold & \CheckmarkBold & \CheckmarkBold & \XSolidBrush &  \mla{78.16} \\
			\CheckmarkBold & \CheckmarkBold & \CheckmarkBold & \CheckmarkBold & \CheckmarkBold & \textbf{78.49} \\
			\bottomrule[1.0pt]
		\end{tabular}
	}%
	\begin{tablenotes}
		\item[1] \textcolor{magenta}{
			\xma{
				* Note that ``Face-Net'' denotes that only Face-Net is used to predict \\ the talking face, and ``spatial'' and ``temporal'' mean adding the com-\\ponent of spatial GAT and temporal GAT in STMG, respectively. \\ Besides, ``audio'' and ``visual'' represent adding multi-modal GAT \\ with audio and visual modality features, respectively.
			}
		} \hspace{-2em}
	\end{tablenotes}
	\label{tab:sound_acc}%
\end{table}%

\noindent\textbf{Visual branch}.
Visual branch leverages basic visual information, \ie, texture, motion, and temporal cues, \xma{and the attention cues from STMG}, to predict saliency.
We evaluate the visual branch of the proposed network and report the results in Tab. \ref{tab:ablation}. 
\xma{It shows that the visual branch using only RGB frames and optical flow maps can reach a CC of 0.682 and KL of 0.933, better than most of other methods and comparable to the second best method TASED. 
	The performance further reaches 0.702 in CC and 0.925 in KL by adding convolutional LSTM to fuse the temporal cues.
	In addition, the utilization of attention weights from the face branch boosts the performance to 0.725 in CC and 0.834 in KL (see ``visual+face'' in Tab. \ref{tab:ablation}).
	This also manifests the effectiveness of the joint learning of saliency prediction and sound source localization.
}
Hence, the entire visual branch and its components are all useful to saliency prediction.
\xma{Moreover, as shown in Tab. \ref{tab:sound_acc}, the combination of face and audio components results in lower performance than combining all components (\ie, the whole network). It further manifests the effectiveness of the visual branch for sound source localization. 
}

\noindent\textbf{Face branch}.
\xma{
	The face branch is designed to localize the sound source and to promote saliency prediction. We first analyze its contribution to saliency prediction. 
	From Tab. \ref{tab:ablation}, the values of CC and KL reach to 0.725 and 0.834, respectively, after integrating the face branch with the visual modality}. 
\xma{
	In other words, the face branch improves the performance of saliency prediction by 0.023 and 0.091 in terms of CC and KL, respectively. This verifies the necessity of incorporating the face branch into our VAM-Net.
	For sound source localization, we analyze the effectiveness of each component in the face branch. 
	As shown in Tab. \ref{tab:sound_acc}, the spatial, temporal, audio, and visual cues all improve the performance of sound source localization. }

\noindent\textbf{Audio branch}.
\xma{In addition to the visual and face branches, we add the audio branch to the framework.}
With the help of the audio branch, \xma{the visual-audio saliency model} achieves \mla{0.741} in CC and \mla{0.783} in KL, much better than the visual branch. 
\xma{
	For sound source localization, as shown in Tab. \ref{tab:sound_acc}, the face branch with STMG embedded audio component has better performance than that without audio component, and it obtains more than 0.9\% Acc improvement.
}
These results manifest the contribution of audio information and the effectiveness of the proposed audio branch.

\begin{figure}[t]
	\begin{center}
		\vspace{1em}
		\includegraphics[width=.99\linewidth]{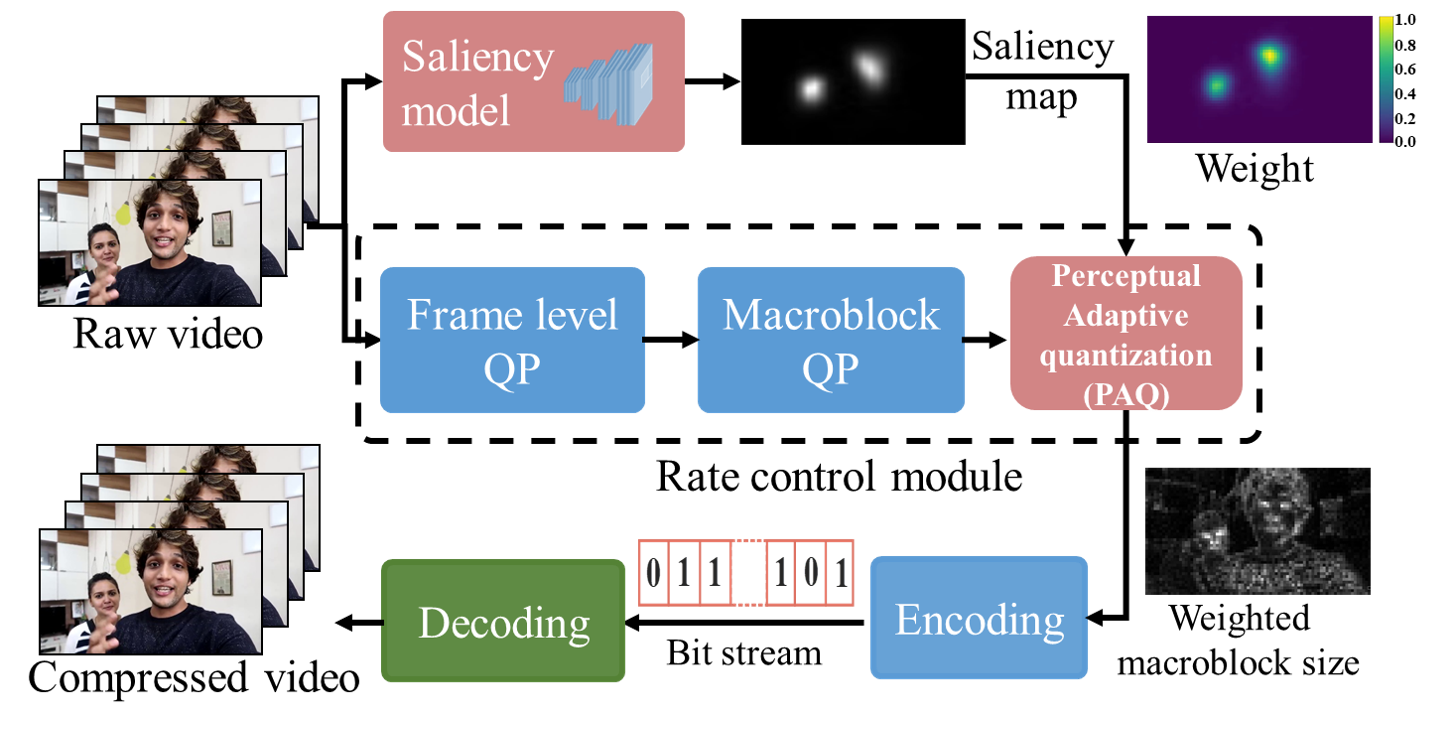}
	\end{center}
	\vspace{-1em}
	\caption{{Our saliency based video perceptual coding framework.}}
	\label{fig:app1}
	\vspace{-1em}
\end{figure}

\noindent\textbf{Baselines}. 
\xma{We evaluate different baselines on the task of saliency prediction for the proposed method.} 
On the one hand, we divide the subjects into 2 groups, and calculate the similarity of these two groups to approximate the human performance. On the other hand, we compute the mean eye position map and regard the assessment as the baseline of average. It can be seen in Tab. \ref{tab:ablation} that our method performs far beyond the baseline of average and reaches close to the human results.

\noindent\xma{\textbf{Joint leaning of two tasks}.
	The proposed method aims to jointly learn the two tasks of saliency prediction and sound source localization. Experiments verify that these two task can boost the performance of each other in our method. For example, as reported in Tab. \ref{tab:ablation}, ``visual+face'' model has better saliency prediction performance than ``visual'' model. That is, ``face'' branch, which takes sound source localization as the main task, is helpful in improving the performance of ``visual'' branch that takes saliency prediction as the main task. 
	Likewise, as shown in Tab. \ref{tab:sound_acc}, the face branch with the help of visual branch (\ie, the fifth row in Tab. \ref{tab:sound_acc}) also performs better than that without the help of visual branch (\ie, the fourth row in Tab. \ref{tab:sound_acc}) for sound source localization. 
	These results indicate that the two tasks of saliency prediction and sound source localization are complementary in our method.
}

In summary, the ablation analysis confirms the necessity of different cues for saliency prediction and sound source localization, and verifies the effectiveness of each part in our model.
\xma{
}
\xma{
	\subsection{Applications}}
\vspace{-4pt}
\xma{
	The proposed saliency prediction method has the potential to be implemented in the video processing tasks. Here, we focus on the application of our saliency prediction method in perceptual video compression. 
	For video compression, our VAM-Net can be utilized to locate salient regions, \ie, regions of interest (ROI), and then perceptual quality of compressed videos can be improved by assigning more coding bits to ROI. The details about our implementation and results are described as follows. 
}

\xma{
	\textbf{Implementation of perceptual video compression.} 
	Our perceptual video method is implemented on the widely used codec, X.264 \citep{x264}. The proposed saliency model was embedded into the rate control (RC) scheme of H.264. The overall framework of our implementation is shown in Fig. \ref{fig:app1}, where blue and  pink blocks distinguish the components of the traditional X.264 codec and our algorithm.
	The saliency maps predicted by our method are fed to the rate control module of X.264.
	Guided by the saliency maps, more bits can be assigned to ROI at a given target bit-rate, via adjusting quantization parameters (QPs) of each coding block.}

\begin{figure*}[htbp]
	\begin{center}
		\vspace{-1.em}
		\includegraphics[width=.9\linewidth]{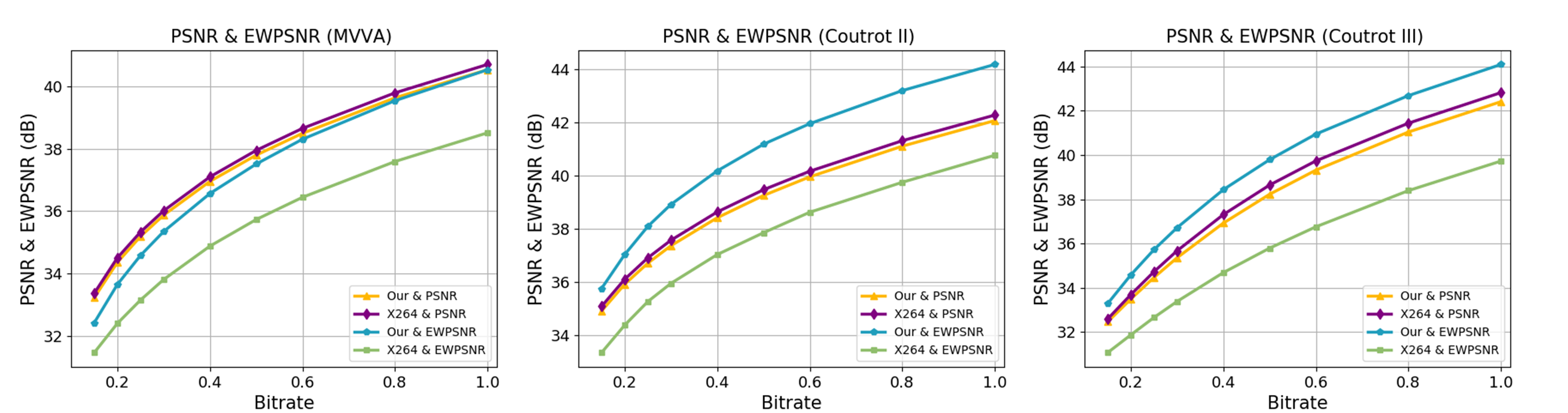}
	\end{center}
	\vspace{-1em}
	\caption{\xma{Rate-distortion curves of our implemented perceptual compression method and the traditional X.264 codec over our MVVA, Coutrot II and Coutrot III databases.}}
	\label{fig:rd_curve}
	
\end{figure*}

\begin{figure*}[htbp]
	\begin{center}
		\vspace{-.5em}
		\includegraphics[width=0.9\linewidth]{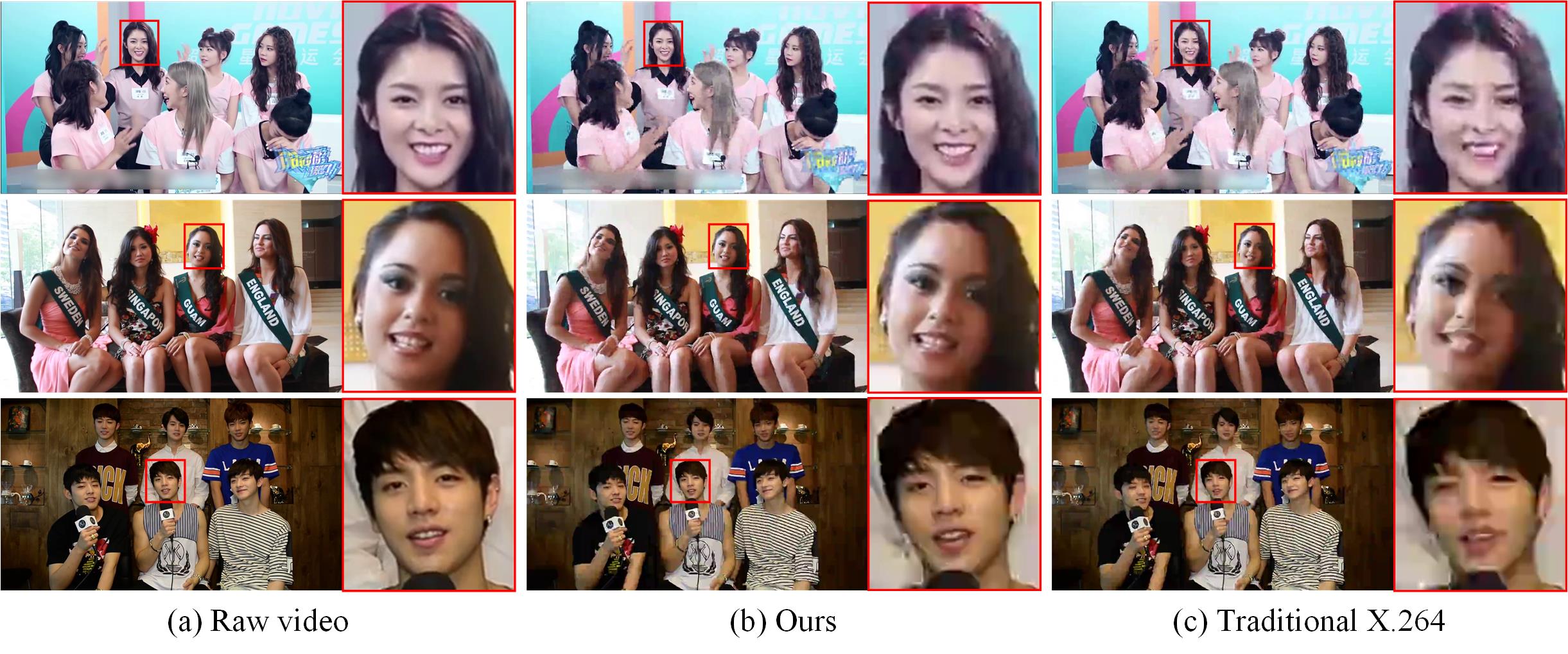}
	\end{center}
	\vspace{-1.5em}
	\caption{\xma{Subjective quality comparison. (b) and (c) are the frames compressed at 650K bits/second  by our perceptual compression method and the traditional X.264 codec, respectively. }}
	\label{fig:app_sub1}
	\vspace{-1em}
\end{figure*}

\yfa{
	\textbf{Results of perceptual video compression.}
	We report the compression results to validate the performance of our implementation. 
	Here, we use eye-tracking weight peak signal to noise ratio (EWPSNR) \citep{li2011visual}, which weights PSNR with human fixation maps, for evaluating the perceptual quality of the compressed video at various bit rates.
	\xma{We compress the test videos at bit-rates of 150, 200, 250, 300, 400, 500, 600, 800 and 1000 kbps.
	}
	\xma{Fig. \ref{fig:rd_curve} compares the PSNR and EWPSNR results of the compressed videos by
		our implementation and the traditional X.264 codec.}
	\xma{As can be seen, our implementation significantly improves the perceptual quality of videos compressed by X.264, \xma{with a gain of 2-3} dB over X.264 in terms of EWPSNR at the same bit-rate.}
	Fig. \ref{fig:app_sub1} further compares the subjective quality. It can be observed that our implementation yields higher quality in ROI (\ie, the salient face), compared with X.264. 
	In summary, our saliency prediction method can be used to improve the perceptual quality of multi-face video compression. 
}

\vspace{4pt}
\section{Conclusion}
In this paper, \xma{we proposed a new method for \xma{simultaneously} predicting visual-audio saliency \xma{and sound source localization} on multi-face videos\xma{, which takes advantage of visual, audio and face information.}} \xma{Specifically, we first introduced the MVVA database which includes fixations of 34 subjects and annotated sound source for 300 multi-face videos.}
Using our database, we then \xma{studied \xma{the factors that} influence human attention
	on multi-face videos. 
	Inspired by our findings, we proposed a novel \xma{visual-audio multi-task network (VAM-Net)} consisting of visual, audio and face branches, for \xma{ the tasks of visual-audio saliency prediction and sound source localization}. The three branches encode visual frames, audio signals and faces into features. Besides, a \xma{spatio-temporal multi-modal graph (STMG)} was designed to integrate the features of the three modalities \xma{and to explore the interaction among multiple faces. We found that joint learning of the tasks of saliency prediction and sound source localization, improves the performance on both tasks.}
	Finally, experimental results showed that our method significantly outperforms 12 state-of-the-art \xma{saliency prediction} methods in terms of 4 metrics, \yfa{and achieves competitive performance on sound source localization}.}

\xma{We foresee three directions for the future research in this area.
	First, it would be interesting to extend our method to visual-audio saliency prediction on generic videos, rather than multi-face videos considered in this paper.
	Second, the acceleration of the proposed method is another promising future work, for making it practical in real-time applications. 
	Third, in addition to perceptual video coding, it is promising to apply our method to other video processing tasks, such as video enhancement and rendering. }

\bibliographystyle{spbasic}
\bibliography{ref}

\end{document}